\documentclass{article} 
\usepackage{preprint,times}

\usepackage[utf8]{inputenc} 
\usepackage[T1]{fontenc}    
\usepackage{url}            
\usepackage{booktabs}       
\usepackage{amsfonts}       
\usepackage{nicefrac}       
\usepackage{microtype}      
\usepackage{xcolor}         

\usepackage{mathtools}
\usepackage{amsmath}
\usepackage{amssymb}
\usepackage{amsthm}
\usepackage{bbm}
\usepackage{dutchcal}
\usepackage{todonotes}
\usepackage[shortlabels]{enumitem}
\usepackage{bm}
\usepackage{bbm}
\usepackage[bookmarks=true, colorlinks=true, linkcolor=blue!50!black,
citecolor=orange!50!black, urlcolor=orange!50!black, pdfencoding=unicode]{hyperref}
\usepackage{wrapfig}
\usepackage{tikz}
\usetikzlibrary{positioning}

\usepackage[capitalize]{cleveref}
\Crefname{equation}{Eq.}{Eqs.}
\Crefname{figure}{Fig.}{Figs.}
\Crefname{table}{Tab.}{Tabs.}
\Crefname{section}{Sec.}{Secs.}
\Crefname{appendix}{App.}{Apps.}
\crefname{lstlisting}{list.}{lists.}
\Crefname{lstlisting}{List.}{Lists.}

\usepackage{xstring}
\usepackage{bookmark}
\usepackage{typed-checklist}
\usepackage{fontawesome}
\usepackage{ulem}
\usepackage{listings}
\theoremstyle{definition}

\newcommand{\XX}{\mathcal{X}}
\newcommand{\RR}{\mathcal{R}}
\newcommand{\DD}{\mathcal{D}}
\newcommand{\PP}{\mathbb{P}}
\newcommand{\LFGSM}{{\ensuremath{\mathcal L\text{-FGSM}}}}
\newcommand{\LPGD}{{\ensuremath{\mathcal L\text{-PGD}}}}
\newcommand{\LBIM}{{\ensuremath{\mathcal L\text{-BIM}}}}
\newcommand{\BLFGSM}{{\ensuremath{\bar{\mathcal L}\text{-FGSM}}}}
\newcommand{\BLPGD}{{\ensuremath{\bar{\mathcal L}\text{-PGD}}}}
\newcommand{\BLBIM}{{\ensuremath{\bar{\mathcal L}\text{-BIM}}}}

\newcommand{\linf}{\ell_\infty}
\DeclareMathOperator{\sign}{sign}
\DeclareMathOperator{\clip}{clip}
\DeclareMathOperator{\highpass}{HighPass}
\DeclareMathOperator{\Sim}{sim}

\title{Robustness of Unsupervised Representation Learning without Labels}

\author{Aleksandar Petrov\\
Department of Engineering Science \\
University of Oxford\\
\texttt{aleks@robots.ox.ac.uk}
\And
Marta Kwiatkowska\\
Department of Computer Science\\
University of Oxford\\
\texttt{marta.kwiatkowska@cs.ox.ac.uk}
}

\begin{document}

\suppressfloats
\maketitle

\begin{abstract}
    Unsupervised representation learning leverages large unlabeled datasets and is competitive with supervised learning.
    But non-robust encoders may affect downstream task robustness.
    Recently, robust representation encoders have become of interest.
    Still, all prior work evaluates robustness using a downstream classification task.
    Instead, we propose a family of unsupervised robustness measures, which are model- and task-agnostic and label-free.
    We benchmark state-of-the-art representation encoders and show that none dominates the rest.
    We offer unsupervised extensions to the FGSM and PGD attacks.
    When used in adversarial training, they improve most unsupervised robustness measures, including certified robustness. 
    We validate our results against a linear probe and show that, for MOCOv2, adversarial training results in 3 times higher certified accuracy, a 2-fold decrease in impersonation attack success rate and considerable improvements in certified robustness.
\end{abstract}

\section{Introduction}

Unsupervised and self-supervised models extract useful representations without requiring labels.
They can learn patterns in the data and are competitive with supervised models for image classification by leveraging large unlabeled datasets \citep{zbontar_barlow_2021,chen_exploring_2021,he_momentum_2020,chen_simple_2020,chen_improved_2020,chen_big_nodate}.
Representation encoders do not use task-specific labels and can be employed for various downstream tasks.
Such reuse is attractive as large datasets can make them expensive to train.

\begin{figure}
    \centering
    \includegraphics[width=0.95\textwidth]{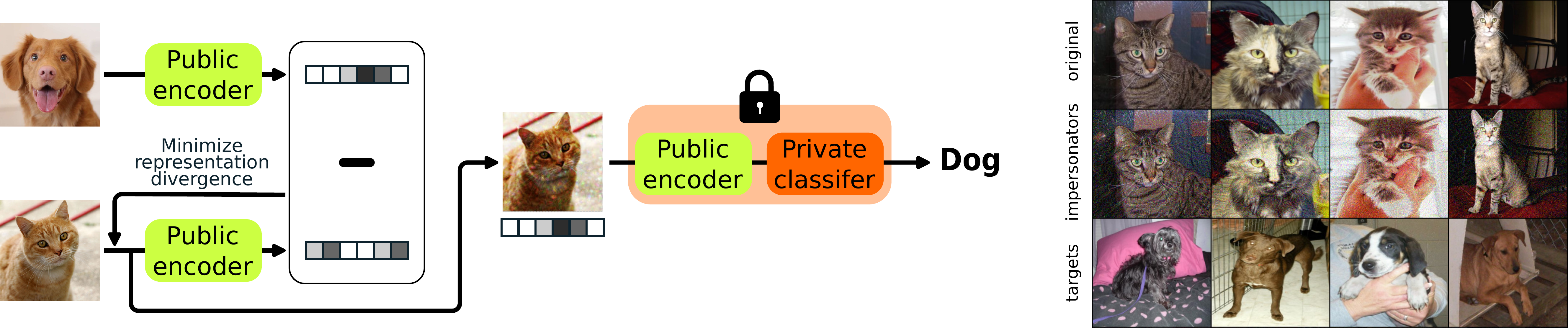}
    \caption{
		Impersonation attack threat model. 
		The attacker has access only to the encoder on which the classifier is built. 
		By attacking the input to have a similar representation to a sample from the target class, the attacker can fool the classifier without requiring any access to it. 
		Cats who successfully impersonate dogs under the MOCOv2 representation encoder and a linear probe classifier are shown.
	}
    \label{fig:threat_model}
\end{figure}

Therefore, applications are often built on top of public domain representation encoders.
However, lack of robustness of the encoder can be propagated to the downstream task.
Consider the impersonation attack threat model in \Cref{fig:threat_model}.
An attacker tries to fool a classifier that uses a representation encoder.
The attacker has white-box access to the representation extractor (e.g.\ an open-source model) but they \emph{do not} have access to the classification model that uses the representations.
By optimizing the input to be similar to a benign input, but to have the representation of a different target input, the attacker can fool the classifier. 
Even if the classifier is private, one can attack the combined system if the public encoder conflates two different concepts onto similar representations.
Hence, robustness against such conflation is necessary to perform downstream inference on robust features.

We currently lack ways to evaluate robustness of representation encoders without specializing for a particular task.
While prior work has proposed improving the robustness of self-supervised representation learning \citep{kim_adversarial_2020,jiang_robust_nodate,ho_contrastive_2020,chen_adversarial_2020,cemgil_adversarially_2020,fan_when_2021,alayrac_are_2019,carmon_unlabeled_2019,nguyen_task_agnostic_2022}, they all require \emph{labeled} datasets to evaluate the robustness of the resulting models.

Instead, we offer encoder robustness evaluation \emph{without labels}. 
This is task-agnostic, in contrast to supervised assessment, as labels are (implicitly) associated with a specific task.
Labels can also be incomplete, misleading or stereotyping \citep{ferrari_convnets_2018,steed_image_2021,birhane_large_2021}, and can inadvertently impose biases in the robustness assessment.
In this work, we propose measures that do not require labels and methods for unsupervised adversarial training that result in more robust models.
To the best of our knowledge, this is the first work on unsupervised robustness evaluation and we make the following contributions to address this problem:
\begin{enumerate}
	\item Novel representational robustness measures based between clean and adversarial representation divergences, requiring no labels or assumptions about underlying decision boundaries.
    \item A unifying framework for unsupervised adversarial attacks and training, which generalizes the prior unsupervised adversarial training methods.
    \item Evidence that even the most basic unsupervised adversarial attacks in the framework result in more robust models relative to both supervised and unsupervised measures. 
	\item Probabilistic guarantees on the unsupervised robustness measures based on center smoothing.
\end{enumerate}

\section{Related work}
\label{sec:related_work}

\paragraph{Adversarial robustness of supervised learning}
 
Deep neural networks can have high accuracy on clean samples while performing poorly under imperceptible perturbations (adversarial examples) \citep{szegedy_intriguing_2014,biggio_evasion_2013}.
Adversarial examples can be viewed as spurious correlations between labels and style \citep{zhang_adversarial_2021,singla_salient_2022} or shortcut solutions \citep{robinson_can_2021}.
Adversarial training, i.e.\ incorporating adversarial examples in the training process, is a simple and widely used strategy against adversarial attacks \citep{goodfellow_explaining_2015,madry_towards_2019,shafahi_adversarial_nodate,bai_recent_2021}.

\paragraph{Unsupervised representation learning}
Representation learning aims to extract useful features from data.
Unsupervised approaches are frequently used to leverage large unlabeled datasets.
Siamese networks map similar samples to similar representations \citep{bromley_signature_nodate,koch_siamese_nodate}, but may collapse to a constant representation.
However, \citet{chen_exploring_2021} showed that a simple stop-grad can prevent such collapse.
Contrastive learning was proposed to address the representational collapse by introducing negative samples \citep{hadsell_dimensionality_2006,le-khac_contrastive_2020}.
It can benefit from pretext tasks \citep{xie_propagate_2021,bachman_learning_2019,tian_contrastive_2019,oord_representation_2019,ozair_wasserstein_2019,mcallester_formal_2019}.
Some methods that do not need negative samples are VAEs \citep{kingma_auto-encoding_2014}, generative models \citep{kingma_semi-supervised_2014,goodfellow_generative_2014,donahue_large_2019}, or bootstrapping methods such as BYOL by \citet{grill_bootstrap_nodate}.

\paragraph{Robustness of unsupervised representation learning}
Most robustness work has focused on supervised tasks, but there has been recent interest in unsupervised training for representation encoders.
\citet{kim_adversarial_2020} and \citet{jiang_robust_nodate} propose generating instance-wise attacks by maximizing a contrastive loss and using them for adversarial training.
\citet{fan_when_2021} complement this by a high-frequency view.
\citet{ho_contrastive_2020} suggest attacking batches instead of individual samples.
KL-divergence can also be used as a loss \citep{alayrac_are_2019} or as a regularizer \citep{nguyen_task_agnostic_2022}.
Alternatively, a classifier can be trained on a small labeled dataset with  adversarial training applied to it \citep{carmon_unlabeled_2019,alayrac_are_2019}.
For VAEs, \citet{cemgil_adversarially_2020} generate attacks by maximizing the Wasserstein distance to the clean representations in representation space.
\citet{peychev_latent_2022} address robustness from the perspective of individual fairness: they certify that samples close in a feature directions are close in representation space. However, their approach is limited to invertible encoders.
While these methods obtain robust unsupervised representation encoders, they all evaluate robustness on a single supervised classification task.
To the best of our knowledge, no prior work has proposed measures for robustness evaluation \emph{without} labels.\footnote{Concurrent work by \citet{wang22_RVCL} proposed RVCL: a method to evaluate robustness without labels. However, it focuses on contrastive learning models while the methods here work with arbitrary encoders.}

\section{Problem setting}
\label{sec:problem_setting}

\begin{wrapfigure}{r}{0.50\textwidth}
	\vspace{-13pt}
	\centering
	\includegraphics[width=0.5\textwidth]{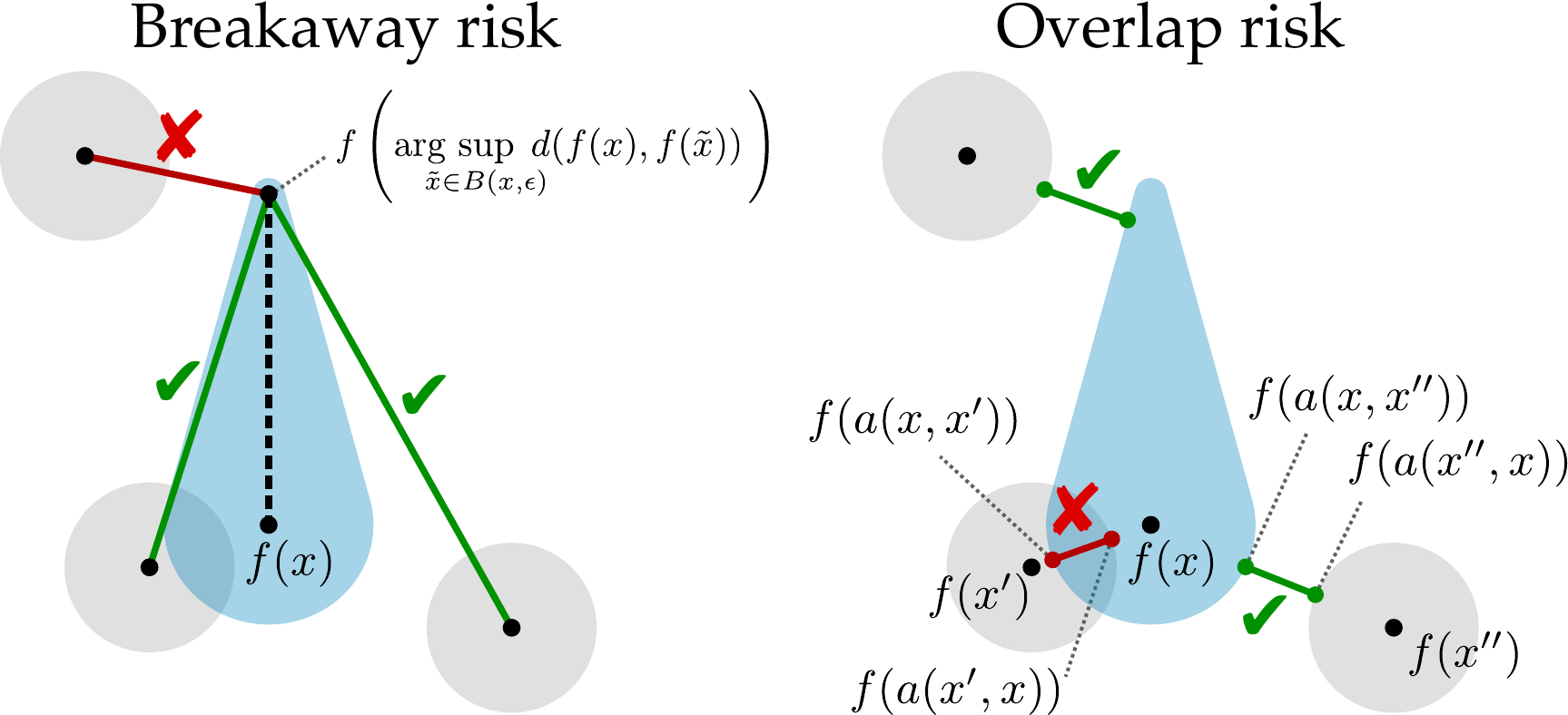}
	\caption{Breakaway and overlap risks. Divergences that increase the corresponding risks are in red and those reducing them are in green.}
	\label{fig:risks}
\end{wrapfigure}
Let $f: \XX \to \RR$ be a differentiable encoder from $\mathcal X= [0,1]^n$ to a representation space $\RR$.
We require $f$ to be white-box: we can query both $f(x)$ and $\frac{df(x)}{dx}$.
$\RR$ is endowed with a divergence $d(r,r')$, a function that has the non-negativity ($d(r,r')\geq 0, \forall r,r' \in \RR$) and identity of indiscernibles ($d(r,r')=0 \Leftrightarrow r=r'$) properties.
This includes metrics on $\RR$ and statistical distances, e.g.\ the KL-divergence.
$D$ is a dataset of iid samples from a distribution $\DD$ over $\XX$.

Perturbations of the input $x$ are denoted as $\hat x = x+\delta, \left\|\delta\right\|_\infty \leq \epsilon$, with $\epsilon$ small enough so that any $\hat x$ is semantically indistinguishable from $x$.
We assume that it is desirable that $f$ maps $\hat x$ close to $x$, i.e.\ $d(f(x),f(\hat x))$ should be ``small''.
We call this property \emph{unsupervised robustness}.
It can also be referred to as \emph{smoothness} \citep{bengio_representation_2013,alayrac_are_2019}, although it is more closely related with the concept of (Lipschitz) continuity.

It is not immediately obvious what value for $d(f(x),f(\hat x))$ would be ``small''.
This is encoder- and input-dependent, as some parts of the representation manifold can be more densely packed than others.
To circumvent this issue, we consider the \emph{breakaway risk}, i.e.\ the probability that the worst-case perturbation of $x$ with maximum size $\epsilon$ is closer to a different sample $x'$ than it is to $x$:
\begin{equation}
    \underset{x,x'\sim\DD}{\PP} \left[ d(f(\hat x), f(x')) < d( f(\hat x), f(x)) \right], ~~ \hat x \in \arg\sup_{\tilde x\in B(x,\epsilon) } d(f(x), f(\tilde x)).
    \label{eq:break_away_risk}
\end{equation}

Another indication of the lack of unsupervised robustness is if $f(B(x,\epsilon))$ and $f(B(x',\epsilon))$ overlap, as then there exist perturbations $\delta,\delta'$ under which $f$ does not distinguish between the two samples, i.e.\ $f(x+\delta)=f(x'+\delta')$.
We call this the \emph{overlap risk} and define it as:
\begin{equation}
    \underset{x,x'\sim\DD}{\PP} \left[ d(f(x), f(a(x',x))) < d(f(x), f(a(x,x'))) \right], ~~ a(o,t) \in \arg\inf_{\tilde x\in B(o,\epsilon) } d(f(t), f(\tilde x)).
    \label{eq:overlap_risk}
\end{equation}

The breakaway risk is based on the perturbation causing the largest divergence in $\RR$, while the overlap risk measures if perturbations are sufficiently concentrated to be separated from other instances (see \Cref{fig:risks}).
Labels are not required: the two risks are defined with respect to the local properties of the representation manifold of $f$ under $\DD$.
In fact, we neither explicitly nor implicitly consider decision boundaries or underlying classes, as we make no assumptions about the downstream task.

What if $x$ and $x'$ are very similar?
Perhaps we shouldn't consider breakaway and overlap in such cases?
We argue against this.
First, the probability of very similar pairs would be low in a sufficiently diverse distribution $\DD$.
Second, there is no clear notion of similarity on $\XX$ without making assumptions about the downstream tasks.
Finally, even if $x$ is very similar to $x'$, it should still be more similar to any $\hat x$ as $x$ and $\hat x$ are defined to be visually indistinguishable. 
We call this the \emph{self-similarity assumption}.

\section{Unsupervised adversarial attacks on representation encoders}
\label{sec:attacks}

It is not tractable to compute the supremum and infimum in \Cref{eq:break_away_risk,eq:overlap_risk} exactly for a general $f$.
Instead, we can approximate them via constrained optimization in the form of adversarial attacks.
This section shows how to modify the FGSM and PGD supervised attacks for these objectives (\Cref{sec:u-fgsm,sec:u-pgd}), as well as how to generalize these methods to arbitrary loss functions (\Cref{sec:loss_attacks}).
The adversarial examples can also be used for adversarial training (\Cref{sec:adversarial_training}).

\subsection{Unsupervised Fast Gradient Sign Method (U-FGSM) attack}
\label{sec:u-fgsm}

The Fast Gradient Sign Method (FGSM) is a popular one-step method to generate adversarial examples in the supervised setting \citep{goodfellow_explaining_2015}.
Its untargeted mode perturbs the input $x$ by taking a step of size $\alpha\in\mathbb R_{>0}$ in the direction of maximizing the classification loss $\mathcal L_\text{cl}$ relative to the true label $y$.
In targeted mode, it minimizes the loss of classifying $x$ as a target class $t\neq y$:
\begin{align*}
    \hat{x} &= \clip (x + \alpha\sign(\nabla_x \mathcal L_\text{cl} (f(x), y))) && \text{untargeted FGSM,} \\
    \hat{x}^{\to t} &= \clip( x - \alpha\sign(\nabla_x \mathcal L_\text{cl} (f(x), t)) ) && \text{targeted FGSM},
\end{align*}
where $\clip(x)$ clips all values of $x$ to be between 0 and 1.

\paragraph{Untargeted U-FGSM}
We can approximate the supremum in \Cref{eq:break_away_risk} by replacing $\mathcal L_\text{cl}$ with the representation divergence $d$, using a small perturbation  $\eta\in\mathbb{R}^n$ to ensure non-zero gradient:
\begin{equation*}
    \hat{x} = \clip( x + \alpha \sign(\nabla_{x} d ( f(x), f(x+\eta) ) )).
\end{equation*}

\citet{ho_contrastive_2020} also propose an untargeted FGSM attack for the unsupervised setting, which requires batches rather than single images and uses a specific loss function, and hence is an instance of the \BLFGSM\ attack in \Cref{sec:loss_attacks}.
The untargeted U-FGSM proposed here is independent of the loss used for training, making it more versatile.

\paragraph{Targeted U-FGSM}
We can also perturb $x_i\in D$ so that its representation becomes close to $f(x_j)$ for some $x_j\in D, x_j\neq x_i$.
Then a downstream model would struggle to distinguish between the attacked input and the target $x_j$.
This approximates the infimum in \Cref{eq:overlap_risk}:
\begin{equation*}
    \hat{x}_i^{\to j} = \clip( x_i - \alpha \sign(\nabla_{x_i} d ( f(x_i), f(x_j) ) )).
\end{equation*}

\subsection{Unsupervised Projected Gradient Descent (U-PGD) attack}
\label{sec:u-pgd}

PGD attack is the gold standard of supervised adversarial attacks \citep{madry_towards_2019}.
It comprises iterating FGSM and projections onto $B(x,\epsilon)$, the $\linf$ ball of radius $\epsilon$ centered at $x$:
\begin{align*}
    \hat x_0 &= \hat x_0^{\to t} = \clip \left( x + U^n[-\epsilon, \epsilon] \right) && \text{randomized initialization,} \\
    \hat x_{u+1} &= \clip( \Pi_{x,\epsilon} [ \hat x_u + \alpha\sign(\nabla_{\hat x_u} \mathcal L_\text{cl} (f(\hat x_u), y)) ]) && \text{untargeted PGD,}  \\
    \hat x_{u+1}^{\to t} &= \clip( \Pi_{x,\epsilon} [ \hat x_u^{\to t} - \alpha\sign(\nabla_{\hat x_u^{\to t}} \mathcal L_\text{cl} (f(\hat x_u^{\to t}), t)) ] ) && \text{targeted PGD.}
\end{align*}

We can construct the unsupervised PGD (U-PGD) attacks similarly to the U-FGSM attack:
\begin{align*}
    \hat x_0 &=  \hat x_0^{\to t} = \clip \left( x + U^n[-\epsilon, \epsilon] \right) && \text{randomized initialization,} \\
    \hat x_{u+1} &= \clip ( \Pi_{x,\epsilon} [ \hat x_u + \alpha\sign(\nabla_{\hat x_u} d(f(\hat x_u), f(x)) ) ] ) && \text{untargeted U-PGD,} \\
    \hat x_{u+1}^{\to t} &= \clip ( \Pi_{x,\epsilon} [ \hat x_u^{\to t} - \alpha\sign(\nabla_{\hat x_u^{\to t}} d( f(\hat x_u^{\to t}), f(x_j))) ] ) && \text{targeted U-PGD.}
\end{align*}

By replacing the randomized initialization with the $\eta$ perturbation in the first iteration of the untargeted case, one obtains an unsupervised version of the BIM attack \citep{kurakin_adversarial_2017}.
The adversarial training methods proposed by \citet{alayrac_are_2019,nguyen_task_agnostic_2022,cemgil_adversarially_2020} can be considered as using U-PGD attacks with specific divergence choices (see \Cref{app:instance_wise_loss_based_attacks_examples}).

\subsection{Loss-based attacks}
\label{sec:loss_attacks}

In both their supervised and unsupervised variants, FGSM and PGD attacks work by perturbing the input in order to maximize or minimize the divergence $d$.
By considering arbitrary differentiable loss functions instead, we can define a more general class of \emph{loss-based attacks}.

\paragraph{Instance-wise loss-based attacks (\LFGSM, \LPGD)}
Given an instance $x\in\XX$ and a loss function $\mathcal L : (\XX\to\RR) \times \XX \to \mathbb R$, the $\LFGSM$ attack takes a step in the direction maximizing $\mathcal L$:
\begin{equation*}
    \hat x = \clip \left( x + \alpha \sign \left( \nabla_x \mathcal L (f, x) \right)\right).
\end{equation*}
Similarly, for a loss function $\mathcal L : (\XX\to\RR) \times \XX \times \XX \to \mathbb R$ taking a representation encoder, a sample, and the previous iteration of the attack, the loss-based PGD attack is defined as:
\begin{align*}
    \hat x_0 &=  \clip \left( x + U^n[-\epsilon, \epsilon] \right), \\
    \hat x_{u+1} &= \clip \left( \Pi_{x,\epsilon} \left[ \hat x_u + \alpha\sign( \nabla_{\hat x_u} \mathcal L(f, x, \hat x_u ) \right] \right).
\end{align*}
If we do not use random initialization for the attack, we get the \LBIM\ attack.

The supervised and unsupervised FGSM and PGD attacks are special cases of the \LFGSM\ and \LPGD\ attacks.
Furthermore, prior unsupervised adversarial training methods can also be represented as \LPGD\ attacks \citep{kim_adversarial_2020,jiang_robust_nodate}.
A full description is provided in \Cref{app:instance_wise_loss_based_attacks_examples}.

\paragraph{Batch-wise loss-based attacks (\BLFGSM, \BLPGD)}
Attacking whole batches instead of single inputs can account for interactions between the individual inputs in a batch.
The above attacks can be naturally extended to work over batches by independently attacking all inputs from $X=[x_1,\ldots,x_N]$.
This can be done with a more general loss function $\bar{\mathcal L} : (\XX\to\RR) \times \XX^N \to \mathbb R$.
The batch-wise loss-based FGSM attack $\BLFGSM$ is provided in \Cref{eq:blfgsm} with $\BLPGD$ and $\BLBIM$ defined similarly.
\begin{equation}
    \hat X = \clip \left( X + \alpha \sign \left( \nabla_X \bar{\mathcal L} (f, X) \right)\right).
	\label{eq:blfgsm}
\end{equation}

Any instance-wise loss-based attack can be trivially represented as a batch-wise attack.
Additionally, prior unsupervised adversarial training methods can also be represented as \BLFGSM\ and \BLPGD\ attacks \citep{ho_contrastive_2020,fan_when_2021,jiang_robust_nodate} (see \Cref{app:batch_wise_loss_based_attacks_examples}).

\subsection{Adversarial training for unsupervised learning}
\label{sec:adversarial_training}

Adversarial training is a min-max problem minimizing a loss relative to a worst-case perturbation that maximizes it \citep{goodfellow_explaining_2015}.
As the worst-case perturbation cannot be computed exactly (similarly to \Cref{eq:break_away_risk,eq:overlap_risk}), adversarial attacks are usually used to approximate it.
Any of the aforementioned attacks can be used for the inner optimization for adversarial training.
Prior works use divergence-based \citep{alayrac_are_2019,cemgil_adversarially_2020,nguyen_task_agnostic_2022} and loss-based attacks \citep{kim_adversarial_2020,jiang_robust_nodate,ho_contrastive_2020,fan_when_2021}.
These methods tend to depend on complex loss functions and might work only for certain models.
Therefore, we propose using targeted or untargeted U-PGD, as well as \BLPGD\ with the loss used for training.
They are simple to implement and can be applied to any representation learning model.
%

\section{Robustness assessment with no labels}
\label{sec:assessing_robustness}

The success of a supervised attack is clear-cut: whether the predicted class is different from the one of the clean sample.
In the unsupervised case, however, it is not clear when an adversarial attack results in a representation that is ``too far'' from the clean one.
In this section, we propose using quantiles to quantify distances and discuss estimating the breakaway and overlap risks (\Cref{eq:break_away_risk,eq:overlap_risk}).

\paragraph{Universal quantiles for untargeted attacks}

For untargeted attacks, we propose measuring $d(f(\hat x), f(x))$ relative to the distribution of divergences between representations of samples from $\DD$.
In particular, we suggest reporting the quantile
$ q = {\PP}_{x',x''\sim\DD} \left[ d (f(x'), f(x'')) \leq d(f(\hat x), f(x)) \right].  $
This measure is independent of downstream tasks and depends only on the properties of the encoder and $\DD$.
We can use it to compare different models, as it is agnostic to the different representation magnitudes models may have.
In practice, the quantile values can be estimated from the dataset $D$.

\paragraph{Relative quantiles for targeted attacks}

Nothing prevents universal quantiles to be applied to targeted attacks.
However, considering that targeted attacks try to ``impersonate'' a particular target sample, we propose using \emph{relative quantiles} to assess their success.
We assess the attack as the distance $d( f(\hat x_i^{\to j}), f(x_j))$ induced by the attack relative to $d(f(x_i), f(x_j))$, the original distance between the clean sample and the target.
The relative quantile for a targeted attack $\hat x_i^{\to j}$ is then the ratio
$ {d(f(\hat x_i^{\to j}), f(x_j))} / {d(f(x_i), f(x_j))}$.

Quantiles are a good way to assess the success of individual attacks or to compare different models.
However, they do not take into account the local properties of the representation manifold, i.e.\ that some regions of $\RR$ might be more densely populated than others.
The breakaway and overlap risk metrics were defined with this exact purpose.
Hence, we propose estimating them.

\paragraph{Estimating the breakaway risk}
While the supremum in \Cref{eq:break_away_risk} cannot be computed explicitly, it can be approximated using the untargeted U-FGSM and U-PGD attacks.
Therefore, we can compute a Monte Carlo estimate of \Cref{eq:break_away_risk} by sampling pairs $(x,x')$ from the dataset $D$ and performing an untargeted attack on $x$, for example with U-PGD.

\paragraph{Nearest neighbour accuracy}
As the breakaway risk can be very small for robust encoders we propose also reporting the fraction of samples in $D'\subseteq D$ whose untargeted attacks $\hat x$ would have their nearest clean neighbour in $D$ being their corresponding clean samples $x$.
That is:
\begin{equation}
    \frac{1}{|D'|} \sum_{x\in D'} \mathbbm{1} \left[ \nexists x'\in D, x'\neq x, \text{ s.t. } d(f(x'),f(\hat x)) < d(f(x),f(\hat x)) \right].
    \label{eq:nearest_neighbour_accuracy}
\end{equation}

\paragraph{Estimating the overlap risk}
The infimums in \Cref{eq:overlap_risk} can be estimated with an unsupervised targeted attack. 
Hence, an estimate of \Cref{eq:overlap_risk} can be computed by sampling pairs $(x_i,x_j)$ from the dataset $D$ and computing the targeted attacks $\hat{x}_i^{\to j}$ and $\hat{x}_j^{\to i}$.
The overlap risk estimate is then the fraction of pairs for which $d(f(x_i), f(\hat{x}_j^{\to i})) < d(f(x_i), f(\hat{x}_i^{\to j}))$.

\paragraph{Adversarial margin}
In \Cref{eq:overlap_risk} one takes into account only whether overlap occurs but not the magnitude of the violation.
Therefore, we also propose looking at the margin between the two attacked representations, normalized by the divergence between the clean samples:
$$ \frac{d(f(x_i), f(\hat x_j^{\to i})) - d(f(x_i), f( \hat x_i^{\to j})) } {d(f(x_i), f(x_j))},$$
for randomly selected pairs $(x_i, x_j)$ from $D$.
If overlap occurs, this ratio would be negative, with more negative values pointing to stronger violations.
The overlap risk is therefore equivalent to the probability of occurrence of a negative adversarial margin.

\paragraph{Certified unsupervised robustness} 
The present work depends on gradient-based attacks, which can be fooled by gradient masking \citep{athalye_obfuscated_2018,uesato_adversarial_2018}.
Hence, we also assess the certified robustness of the encoder.
By using center smoothing \citep{kumar_center_2021} we can compute a probabilistic guarantee on the radius of the $\ell_2$-ball in $\RR$ that contains at least half of the probability mass of $f(x+\mathcal N(0, \sigma^2))$.
The smaller this radius is, the closer $f$ maps similar inputs.
Hence, this is a probabilistically certified alternative to assessing robustness via untargeted attacks. 
In order to compare certified radius values in $\RR$ across models we report them as universal quantiles.

\section{Experiments}
\label{sec:experiments}

We assess the robustness of state-of-the-art representation encoders against the unsupervised attacks and robustness measures outlined in \Cref{sec:attacks,sec:assessing_robustness}.
We consider the ResNet50-based self-supervised learning models MOCOv2 (200 epochs) \citep{he_momentum_2020,chen_improved_2020}, MOCO with non-semantic negatives (+Patch, $k$=16384, $\alpha$=3) \citep{ge_robust_nodate}, PixPro (400 epochs) \citep{xie_propagate_2021}, AMDIM (Medium) \citep{bachman_learning_2019}, SimCLRv2 (depth 50, width 1x, without selective kernels) \citep{chen_big_nodate}, and SimSiam (100 epochs, batch size 256) \citep{chen_exploring_2021}.
To compare the self-supervised and the supervised methods, we also evaluate the penultimate layer of ResNet50 \citep{he_deep_2016}. 
We assess the effect of using different unsupervised attacks by fine-tuning MOCOv2 with the untargeted U-PGD, targeted U-PGD, as well as with \BLPGD\ using MOCOv2's contrastive loss, as proposed in \Cref{sec:adversarial_training}. 
See \Cref{app:moco2_training} for details and pseudocode.
Additionally, in \Cref{app:extended_results} we evaluate the transformer-based models MAE \citep{he_masked_2021} and MOCOv3 \citep{chen_empirical_2021} as well as adversarially fine-tuned versions of MOCOv3 using the same three attacks as for MOCOv2.

The unsupervised evaluation uses the PASS dataset as it does not contain people and identifiable information and has proper licensing \citep{asano_pass_2021}.
ImageNet \citep{russakovsky_imagenet_2015} is used for accuracy benchmarking and the adversarial fine-tuning of MOCO, as to be consistent with how the model was trained.
Assira \citep{elson_asirra_2007} is used for the impersonation attacks.

We report median universal and relative quantiles for the $\ell_2$ divergence for respectively untargeted and targeted U-PGD attacks with $\epsilon=0.05$ and $\epsilon=0.10$.
In \Cref{app:extended_results} we also report the median $\linf$ divergence and cosine similarity.
We also estimate the breakaway risk, nearest neighbour accuracy, overlap risk and adversarial margin, and certified unsupervised robustness, as described in \Cref{sec:assessing_robustness}.

As customary, we measure the quality of the representations with the top-1 and top-5 accuracy of a linear probe.
We also report the accuracy on samples without high-frequency components, as models might be overly reliant on the high-frequency features in the data \citep{wang_high-frequency_2020}.
Additionally, we assess the certified robustness via randomized smoothing \citep{cohen_certied_2019} and report the resulting Average Certified Radius \citep{Zhai2020MACER}.

In line with the impersonation threat model, we also evaluate to what extent attacking a representation encoder can fool a private downstream classifier.
Pairs of cats and dogs from the Assira dataset \citep{elson_asirra_2007} are attacked with targeted U-PGD so that the representation of one is close to the other.
We report the percentage of impersonations that successfully fool the linear probe.

All implementation details for these experiments can be found in \Cref{app:implementation_details}.

\section{Results}
\label{sec:results}

In this section, we present the results of the experiments on ResNet50 and the ResNet50-based unsupervised encoders.
We defer the results for transformer architectures to \Cref{app:extended_results}.

\begin{table}
    \caption{Standard and lowpass accuracy of linear probes of ResNet50-based encoders.}
    \label{tab:results_resnet_accuracy}
	\centering
    \includegraphics[scale=0.6]{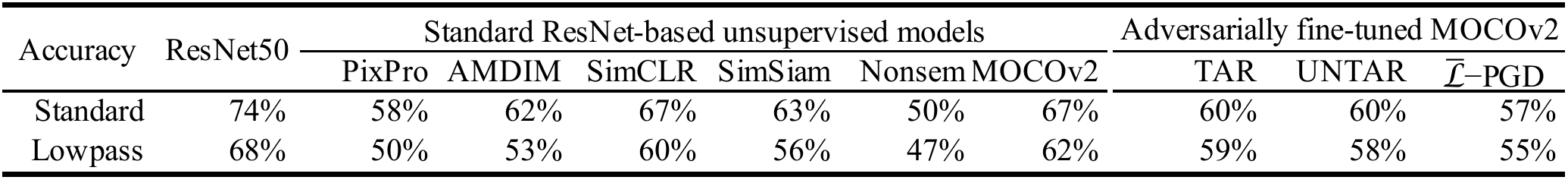}
\end{table}

\begin{table}
	\caption{Robustness of ResNet50 and ResNet50-based unsupervised encoders without unsupervised adversarial training. Arrows show if larger or smaller values are better.}
    \label{tab:results_resnet_robustness}
	\centering
    \includegraphics[scale=0.6]{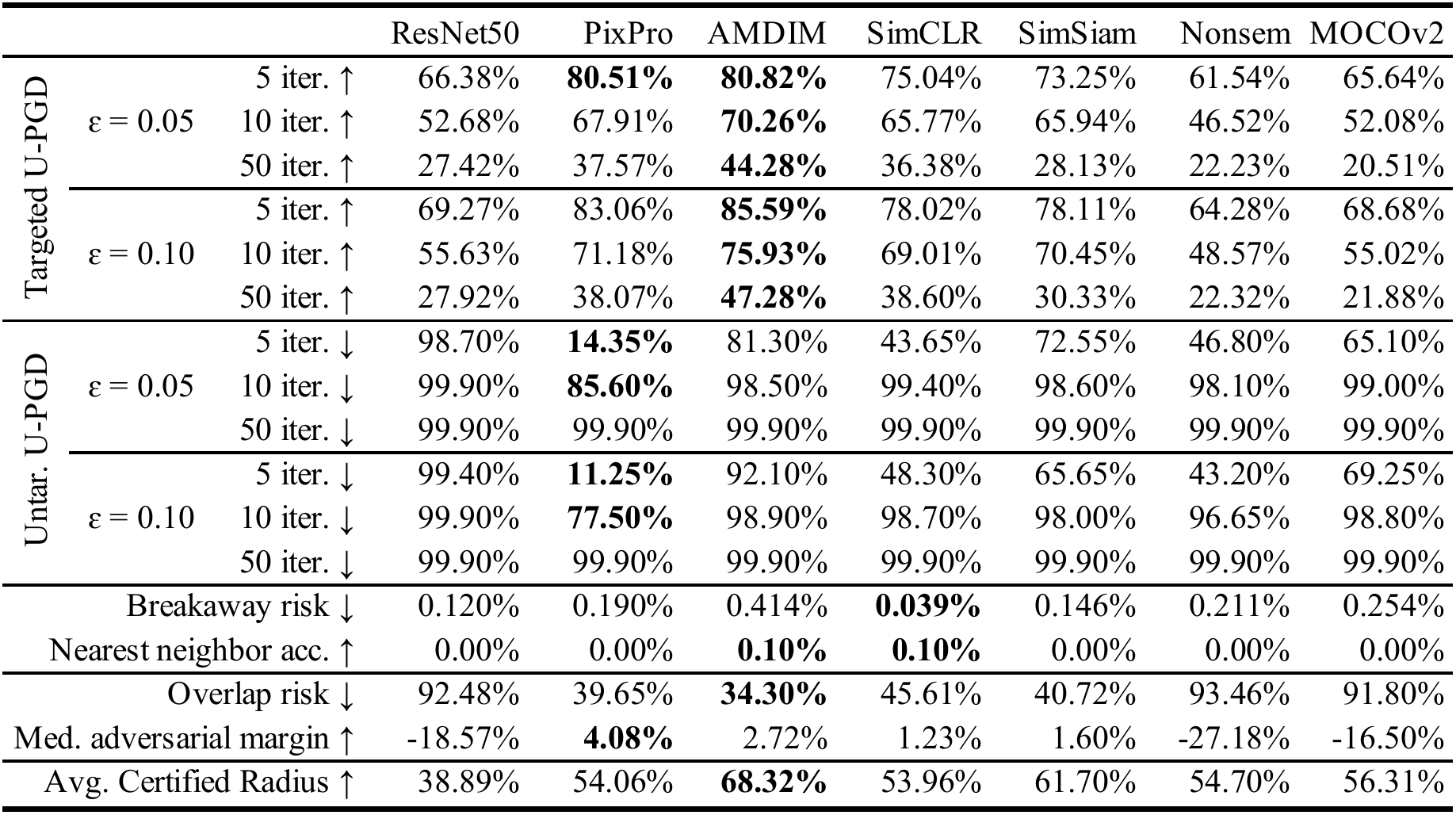}
\end{table}

\paragraph{There is no ``most robust'' standard model.}
Amongst the standard unsupervised models, none dominates on all unsupervised robustness measures (see \Cref{tab:results_resnet_robustness}).
AMDIM is least susceptible to targeted U-PGD attacks and has the highest average certified radius but has the worst untargeted U-PGD, breakaway risk and nearest neighbor accuracy.
PixPro significantly outperforms the other models on untargeted attacks.
AMDIM and PixPro also have the lowest overlap risk and largest median adversarial margin.
At the same time, the model with the lowest breakaway risk is SimCLR. 
While either AMDIM or PixPro scores the best at most measures, they both have significantly higher breakaway risk than SimCLR.
Therefore, no model is a clear choice for the ``most robust model''.

\paragraph{Unsupervised robustness measures reveal significant differences among standard models.}
%
The gap between the best and worst performing unsupervised models for the six measures based on targeted U-PGD attacks is between 19\% and 27\%.
The gap reaches almost 81\% for the untargeted case (PixPro vs AMDIM, 5 it.), demonstrating that standard models on both extremes do exist.
AMDIM has 10.5 times higher breakaway risk than SimCLR while at the same time 2.7 times lower overlap risk than MOCOv2.
Observing values on both extremes of all unsupervised robustness metrics testifies to them being useful for differentiating between the different models.
Additionally, AMDIM having the highest breakaway risk and lowest overlap risk indicates that unsupervised robustness is a multifaceted problem and that models should be evaluated against an array of measures.


\begin{wraptable}{r}{0.54\textwidth}
	\vspace{-20pt}
	\caption{Robustness of MOCOv2 and its adversarially fine-tuned versions. Arrows show if larger or smaller values are better.}
    \label{tab:results_advtrained_resnet_robustness}
	\centering
    \includegraphics[scale=0.6]{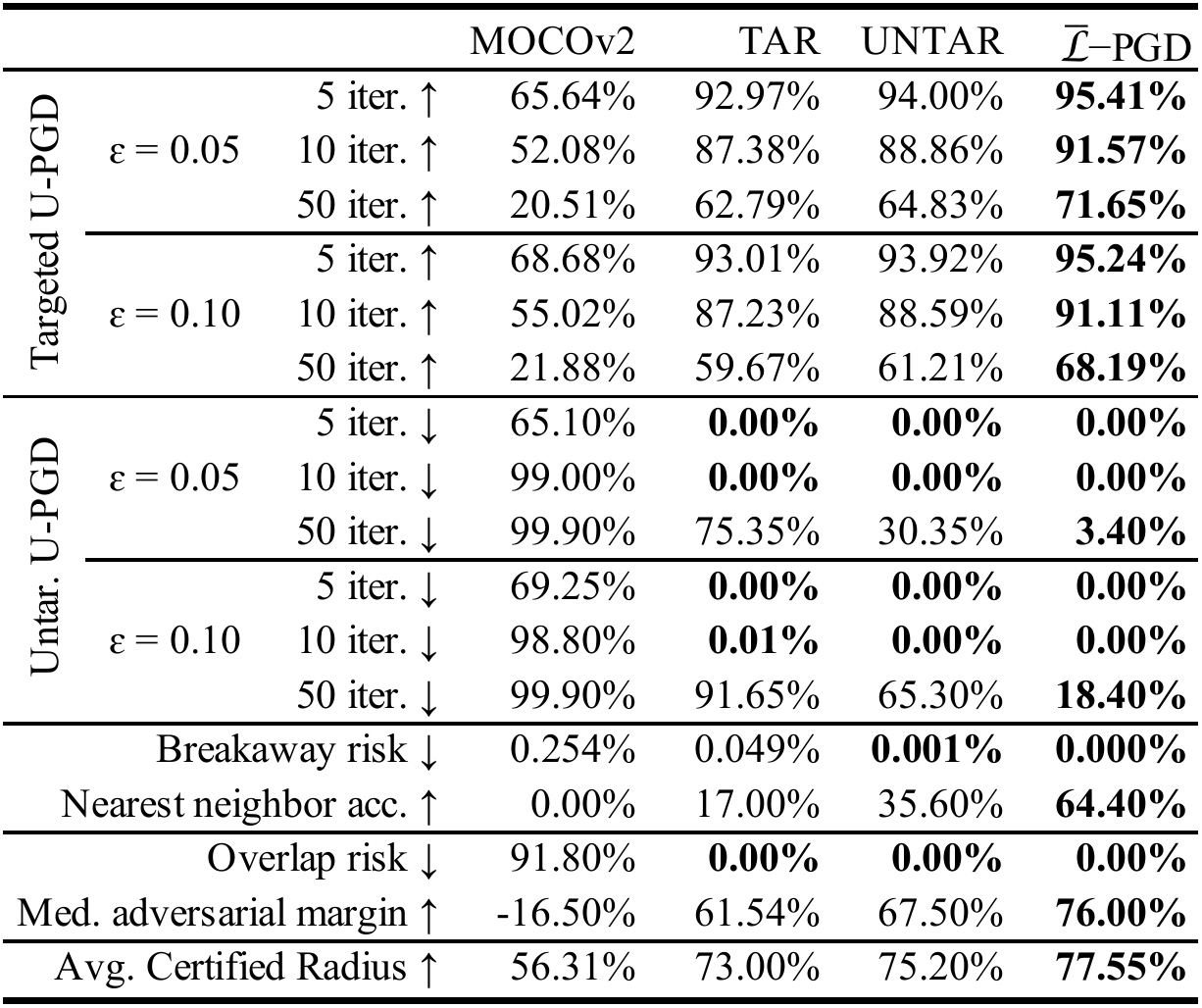}
\end{wraptable}

\begin{figure}
    \centering
    \begin{minipage}[t]{0.49\textwidth}
        \centering
        \includegraphics[width=0.99\textwidth]{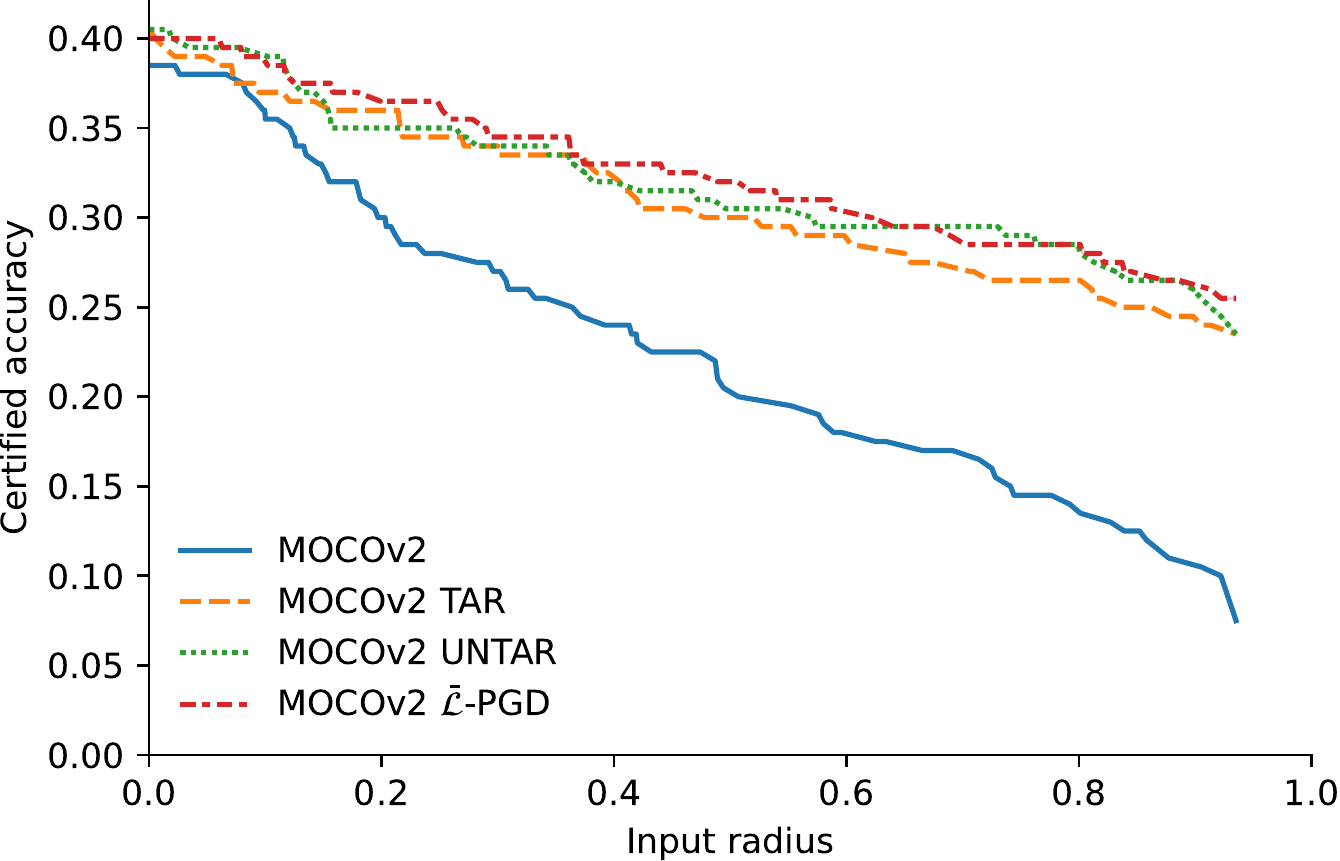}
        \caption{Certified accuracy of randomized smoothed MOCOv2 and its adversarially trained variants on ImageNet. The adversarially trained models are uniformly more robust.}
        \label{fig:certified_accuracy}
    \end{minipage}\hfill
    \begin{minipage}[t]{0.49\textwidth}
        \centering
        \includegraphics[width=0.99\textwidth]{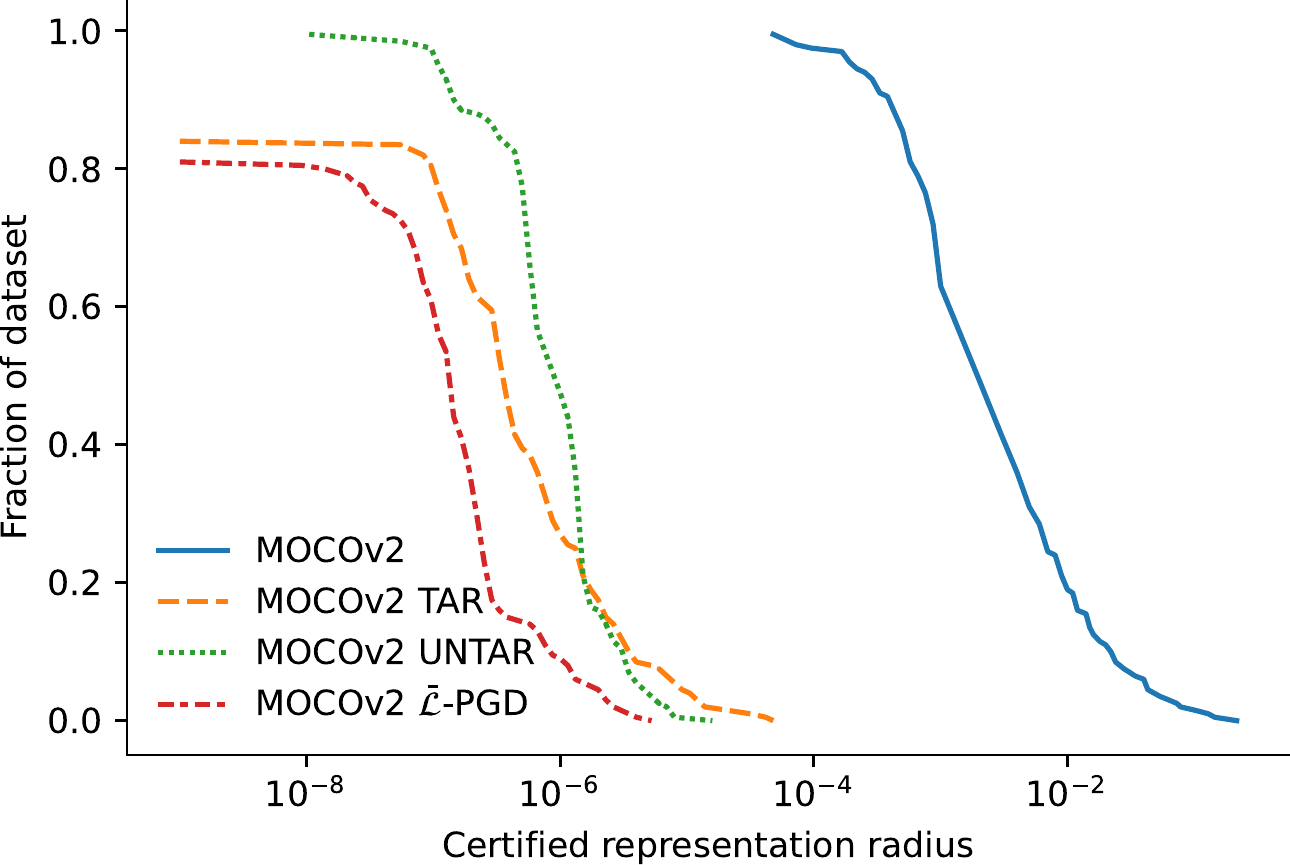}
        \caption{Certified robustness of MOCOv2 on PASS using center smoothing. The certified representation radius is represented as percentile of the distribution of clean representation distances. }
        \label{fig:certified_robustness}
    \end{minipage}
\end{figure}

\paragraph{Unsupervised adversarial training boosts robustness across all measures.}
Across every single unsupervised measure, the worst adversarially trained model performs better than the best standard model (\Cref{tab:results_resnet_robustness,tab:results_advtrained_resnet_robustness}).
Comparing the adversarially trained models with MOCOv2, we observe a significant improvement across the board (\Cref{tab:results_advtrained_resnet_robustness}).
They are also more certifiably robust (\Cref{fig:certified_robustness}).
However, the added robustness comes at the price of reduced accuracy (7\% to 10\%, \Cref{tab:results_resnet_accuracy}), as is typical for adversarial training \citep{zhang_theoretically_2019,tsipras_robustness_2019}.
This gap can likely be reduced by fine-tuning the trade-off between the adversarial and standard objectives and by having separate batch normalization parameters for standard and adversarial samples \citep{kim_adversarial_2020,ho_contrastive_2020}.
Adversarial training also reduces the impersonation rate of a downstream classifier at 5 iterations by  a half relative to MOCOv2 (\Cref{tab:results_advtrained_resnet_impersonation}).
For 50 iterations, the rate is similar to MOCOv2 but the attacked images of the adversarially trained models have stronger semantically meaningful distortions, which can be detected by a human auditor (see \Cref{app:impersonation_attacks_samples} for examples).
These results are for only 10 iterations of fine-tuning of a standard-trained encoder.
Further impersonation rate reduction can likely be achieved with adversarial training applied to the whole 200 epochs of training. 

\begin{wraptable}{r}{0.45\textwidth}
	\vspace{-10pt}
	\caption{Impersonation attack success rate on MOCOv2 and its label-free adversarially fine-tuned versions for different attack iterations.}
    \label{tab:results_advtrained_resnet_impersonation}
	\centering
    \includegraphics[scale=0.6]{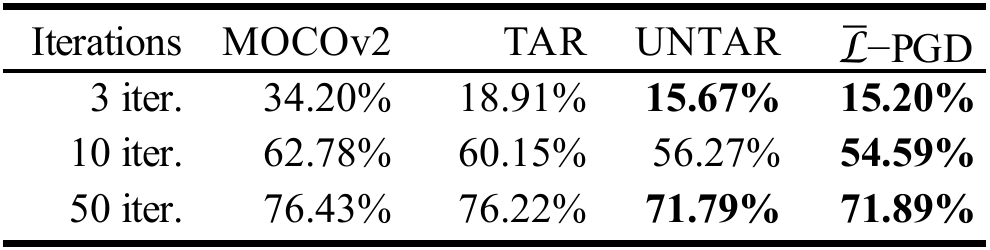}
\end{wraptable}

\paragraph{Unsupervised adversarial training results in certifiably more robust classifiers.}
\Cref{fig:certified_accuracy} shows how the randomized smoothened linear probes of the adversarially trained models uniformly outperform MOCOv2.
The difference is especially evident for large radii: 3 times higher certified accuracy when considering perturbations with radius of 0.935.
These results demonstrate that unsupervised adversarial training boosts the downstream certified accuracy.

\paragraph{Adversarially trained models have better consistency between standard and low-pass accuracy.}
The difference between standard and low-pass accuracy for the adversarially trained models is between 1.7\% and 2.1\%, compared to 2.2\% to 9.7\% for the standard models (\Cref{tab:results_resnet_accuracy}).
This could be in part due to the lower accuracy of the adversarially trained models.
However, compared with PixPro, AMDIM and SimSiam, which have similar accuracy but larger gaps, indicate that the lower accuracy cannot fully explain the lower gap.
Therefore, this suggests that unsupervised adversarial training can help with learning the robust low-frequency features and rejecting high-frequency non-semantic ones.

\paragraph{\BLPGD\ is the overall most robust model, albeit with lower accuracy.}
\BLPGD\ dominates across all unsupervised robustness measures.
These results support the findings of prior work on unsupervised adversarial training using batch loss optimization \citep{ho_contrastive_2020,fan_when_2021,jiang_robust_nodate}.  
However, \BLPGD\ also has lower certified accuracy for small radius values and the lowest supervised accuracy of the three models, as expected due to the accuracy-robustness trade-off.
Still, the differences between the three models are small, and hence all three adversarial training methods can improve the robustness of unsupervised representation learning models.

\section{Discussion, limitations and conclusion}

Unsupervised task-independent adversarial training with simple extensions to classic adversarial attacks can improve the robustness of encoders used for multiple downstream tasks, especially when released publicly.
That is why we will release the adversarially fine-tuned versions of MOCOv2 and MOCOv3,  which can be used as more robust drop-in replacements for applications built on top of these two models.
We showed how to assess the robustness of such encoders without resorting to labeled datasets or proxy tasks.
However, there is no single ``unsupervised robustness measure'': models can have drastically different performance across the different metrics.
Still, unsupervised robustness is a stronger requirement than classification robustness as it requires not only the output but also an intermediate state of the model to not be sensitive to small perturbations. 
Hence, we recommend unsupervised assessment and adversarial training to also be applied to supervised tasks.

We do not compare with the prior methods in \Cref{sec:related_work} as different base models, datasets and objective functions hinder a fair comparison.
Moreover, the methods we propose generalize the previous works, hence this paper strengthens their conclusions rather than claiming improvement over them.

This work is not an exhaustive exploration of unsupervised attacks, robustness measures and defences.
We adversarially fine-tuned only two models, one based on ResNet50 and one transformer-based (in \Cref{app:extended_results});  assessing how these techniques work on other architectures is further required.
There are many other areas warranting further investigation, such as non-gradient based attacks, measures which better predict the robustness of downstream tasks, certified defences, as well as studying the accuracy-robustness trade-off for representation learning.
Still, we believe that robustness evaluation of representation learning models is necessary for a comprehensive assessment of their performance and robustness.
This is especially important for encoders used for applications susceptible to impersonation attacks.
Therefore, we recommend reporting unsupervised robustness measures together with standard and low-pass linear probe accuracy when proposing new unsupervised and supervised learning models.
We hope that this paper illustrates the breadth of opportunities for robustness evaluation in representation space and inspires further work on it.

\ificlrfinal
	\section*{Acknowledgements}
	AP is supported by a grant from Toyota Motor Europe. 
	MK received funding from the ERC under the EU’s Horizon 2020 research and innovation programme (FUN2MODEL, grant No. 834115).
\fi

\section*{Ethics Statement}

This work discusses adversarial vulnerabilities in unsupervised models and therefore exposes potential attack vectors for malicious actors.
However, it also proposes defence strategies in the form of adversarial training which can alleviate the problem, as well as measures to assess how vulnerable representation learning models are.
Therefore, we believe that it would empower the developers of safety-, reliability- and fairness-critical systems to develop safer models.
Moreover, we hope that this work inspires further research into unsupervised robustness,  which can contribute to more robust and secure machine learning systems.

\section*{Reproducibility Statement}

The experiments in this paper were implemented using open-source software packages \citep{harris_array_2020,virtanen_scipy_2020,mckinney_data_2010,paszke_pytorch_2019}, as well as the publicly available MOCOv2 and MOCOv3 models \citep{he_momentum_2020,chen_improved_2020,chen_empirical_2021}.
We provide the code used for the adversarial training, as well as all the robustness evaluation implementations, together with documentation on their use.
We also release the weights of the models and linear probes.
The code reproducing all the experiments in this paper is provided as well.
\ificlrfinal
The details are available \href{https://github.com/AleksandarPetrov/unsupervised-robustness}{here}.
\else
The details are available \href{https://drive.google.com/file/d/15vafFw2My4ll3KRrzyRHfq8KNcsThsnI/view?usp=sharing}{here}.
\fi

\vspace{1cm}

\bibliography{bibliography.bib}
\bibliographystyle{iclr2023_conference}

\newpage

\appendix

\section{Examples of loss-based attacks}
\label{app:loss_based_attacks_examples}

\begin{figure}
    \resizebox{\textwidth}{!}{
        \begin{tikzpicture}[node distance=2cm,line width=1pt]
            \node(BLFGSM)           at (0,0)                                                            {\BLFGSM};
            \node(LFGSM)            [below =1cm and 0cm  of BLFGSM, text width=2cm,align=center]        {\LFGSM};
            \node(HoVasc)           [below right  =1cm and 0cm of BLFGSM, text width=3cm,align=center]  {\citep{ho_contrastive_2020}};
            \node(untarUFGSM)       [below =1cm and 0cm of LFGSM, text width=2cm,align=center]          {Untargeted\\U-FGSM};
            \node(tarUFGSM)         [below right  =1cm and 0cm of LFGSM, text width=2cm,align=center]   {Targeted\\U-FGSM};
            \node(untarFGSM)        [below =1cm and 0cm  of untarUFGSM, text width=2cm,align=center]    {Untargeted\\FGSM};
            \node(tarFGSM)          [below =1cm and 0cm  of tarUFGSM, text width=2cm,align=center]      {Targeted\\FGSM};
            
            \node(BLPGD)            [right = 7cm of BLFGSM, text width=2cm,align=center]                {\BLPGD};
            \node(JiangAtADS)       [below left  =1cm and 0.2cm of BLPGD, text width=3cm,align=center]  {AtA, DS \\\citep{jiang_robust_nodate}};
            \node(LPGD)             [below =1cm and 0.2cm of BLPGD, text width=2cm,align=center]        {\LPGD};
            \node(Fan)              [below right  =1cm and 0.2cm of BLPGD, text width=2cm,align=center] {\citep{fan_when_2021}};
            \node(Kim)              [below left  =1cm and 1.1cm of LPGD, text width=2cm,align=center]   {\citep{kim_adversarial_2020}};
            \node(untarUPGD)        [below left  =1cm and -1.2cm of LPGD, text width=2cm,align=center]  {Untargeted\\U-PGD};
            \node(tarUPGD)          [below right  =1cm and 1.8cm of LPGD, text width=2cm,align=center]  {Targeted\\U-PGD};
            \node(JiangAtS)         [below right  =1cm and -1.4cm of LPGD, text width=3cm,align=center] {AtS\\\citep{jiang_robust_nodate}};
            \node(tarPGD)           [below of=tarUPGD, text width=2cm,align=center] {Targeted\\PGD};
            \node(untarPGD)         [below left =1cm and 1cm of untarUPGD, text width=2cm,align=center] {Untargeted\\PGD};
            \node(Cemgil)           [below left =1cm and -1cm of untarUPGD, text width=2cm,align=center]{\citep{cemgil_adversarially_2020}};
            \node(Alayrac)          [below right =1cm and -1cm of untarUPGD, text width=2cm,align=center]{\citep{alayrac_are_2019}};
            \node(Nguyen)           [below right =1cm and 1cm of untarUPGD, text width=2cm,align=center]{\citep{nguyen_task_agnostic_2022}};
    
            \draw(BLFGSM)           -- (LFGSM);
            \draw(BLFGSM)           -- (HoVasc);
            \draw(LFGSM)            -- (untarUFGSM);
            \draw(LFGSM)            -- (tarUFGSM);
            \draw(untarUFGSM)       -- (untarFGSM);
            \draw(tarUFGSM)         -- (tarFGSM);
    
            \draw(BLPGD)            -- (LPGD);
            \draw(BLPGD)            -- (JiangAtADS);
            \draw(BLPGD)            -- (Fan);
            \draw(LPGD)             -- (Kim);
            \draw(LPGD)             -- (JiangAtS);
            \draw(LPGD)             -- (untarUPGD);
            \draw(LPGD)             -- (tarUPGD);
            \draw(tarUPGD)          -- (tarPGD);
            \draw(untarUPGD)        -- (untarPGD);
            \draw(untarUPGD)        -- (Alayrac);
            \draw(untarUPGD)        -- (Cemgil);
            \draw(untarUPGD)        -- (Nguyen);
    
        \end{tikzpicture}
    }
    \caption{The hierarchy of supervised and unsupervised attacks.}
    \label{fig:attacks_hierarchy}
\end{figure}
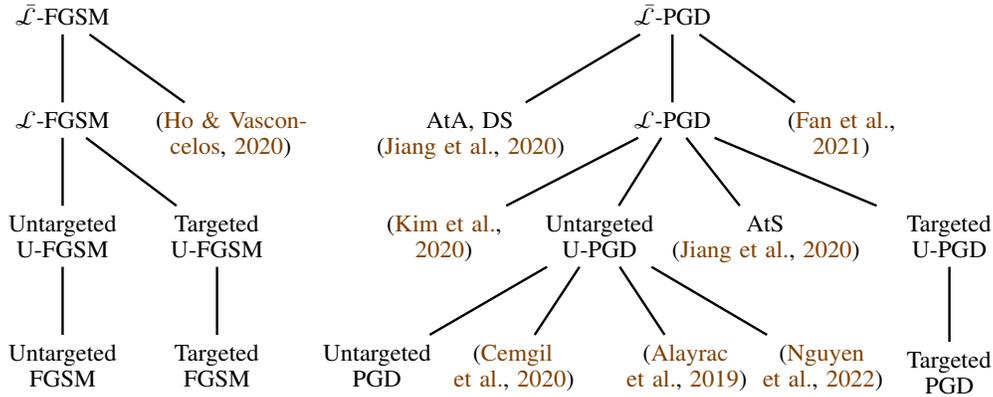

In this appendix we demonstrate how various supervised and unsupervised attacks can be represented as loss-based attacks.
This generalizes the unifying view presented by \citet{madry_towards_2019} by incorporating also unsupervised attacks.
We also illustrate the resulting structure of which attacks are generalizations of other attacks in \Cref{fig:attacks_hierarchy}.
In the following, $\pi(x)$ is the true class of the instance $x$, $\bar{\pi}(x)$ is any other class, $\sigma(x)$ is a function that returns a sample from $\DD$ such that $\sigma(x)\neq x$, and $\kappa(x)$ provides a different view of $x$, e.g.\ a different augmentation.

\subsection{Examples of instance-wise loss-based attacks (\LFGSM, \LPGD, \LBIM)}
\label{app:instance_wise_loss_based_attacks_examples}

\begin{itemize}

    \item The supervised FGSM and PGD attacks can be considered as special cases of the unsupervised $\LFGSM$ and $\LPGD$, where $f$ is a classifier rather than an encoder and we use a classification loss:
        \begin{itemize}
            \item Untargeted FGSM attack: \LFGSM\ with  $\mathcal L (f, x) = \mathcal L_\text{cl} (f(x), \pi(x))$,
            \item Targeted FGSM attack: \LFGSM\ with $\mathcal L (f, x) = -\mathcal L_\text{cl} (f(x), \bar{\pi}(x))$,
            \item Untargeted PGD attack: \LPGD\ with  $\mathcal L (f, x, \hat x_u) = \mathcal L_\text{cl} (f(\hat x_u), \pi(x))$,
            \item Targeted PGD attack: \LPGD\ with $\mathcal L (f, x, \hat x_u) = -\mathcal L_\text{cl} (f(\hat x_u), \bar{\pi}(x))$.
        \end{itemize}

    \item The U-LGSM and U-PGD attacks can be represented as $\LFGSM$ and $\LPGD$ attacks where the loss is the divergence $d$:
        \begin{itemize}
            \item Untargeted U-FGSM attack: \LFGSM\ with $\mathcal{L} (f,x) = d(f(x), f(x+\eta))$,
            \item Targeted U-FGSM attack: \LFGSM\ with $\mathcal{L} (f,x) = -d (f(x), f(\sigma(x)))$,
            \item Untargeted U-PGD attack: \LPGD\ with $\mathcal{L} (f,x,\hat x_u) = d(f(\hat x_u), f(x))$,
            \item Targeted U-PGD attack: \LPGD\ with $\mathcal{L} (f,x,\hat x_u) = -d (f(\hat x_u), f(\sigma(x)))$.

        \end{itemize}

    \item Untargeted U-PGD with the Kullback-Leibler divergence corresponds to the adversarial example generation process of UAT-OT \citep{alayrac_are_2019} which is based on the Virtual Adversarial Training method for semi-supervised adversarial learning \citep{miyato_virtual_2018}.
        Untargeted U-PGD with the Kullback-Leibler also corresponds to the robustness regularizer proposed by \citet{nguyen_task_agnostic_2022}.

    \item \citet{cemgil_adversarially_2020} propose an unsupervised attack for Variational Auto-Encoders (VAEs) \citep{kingma_auto-encoding_2014} based on the Wasserstein distance.
    It can be represented as the untargeted U-PGD attack with the Wasserstein distance, or equivalently, as the \LPGD\ attack with the loss
    \begin{equation*}
        \mathcal L (f, x, \hat x_u) = \mathcal W \left( \mathcal{N}([f(x)]_{\bm{\mu}}, \bm{I} [f(x)]_{\bm\sigma}) , ~ \mathcal{N}[(f(\hat x_u)]_{\bm\mu}, \bm{I} [f(\hat x_u)]_{\bm\sigma}) \right),
    \end{equation*}
    where $\mathcal W$ is the Wasserstein distance, $\mathcal N$ is the normal distribution, $\bm{I}$ is the identity matrix and the subscripts $\bm\mu$ and $\bm\sigma$ designate the respective outputs of the VAE encoder $f$.

    \item The instance-wise unsupervised adversarial attack proposed by \citet{kim_adversarial_2020} is equivalent to \LPGD\ with the contrastive loss
        \begin{equation*}
            \mathcal L (f, x, \hat x_u) = -\log \frac
            { \exp\left( f(\hat x_u)^\top f(\kappa(x)) / T \right) }
            { \exp\left( f(\hat x_u)^\top f(\kappa(x)) / T \right) +  \exp\left( f(\hat x_u)^\top f(\sigma(x)) / T \right)  },
        \end{equation*}
        where $T$ is a temperature parameter.
    This loss encourages that the cosine similarity between the adversarial example and another view of the same sample is small relative to the cosine similarity between the adversarial example and another sample from $\DD$.

    \item Using the NT-Xent loss with \LPGD\ results in the attack used for the Adversarial-to-Standard adversarial contrastive learning proposed by \citet{jiang_robust_nodate}.

\end{itemize}

\subsection{Examples of batch-wise loss-based attacks (\BLFGSM, \BLPGD, \BLBIM)}
\label{app:batch_wise_loss_based_attacks_examples}

\begin{itemize}

    \item Any \LFGSM\ attack with loss $\mathcal L$ is trivially an \BLFGSM\ attack by taking $N=1$ and considering the loss function $\bar{\mathcal L} = \mathcal L$.

    \item Similarly, any \LPGD\ attack is trivially an \BLPGD\ attack.

    \item The adversarial attack proposed by \citet{ho_contrastive_2020} is equivalent to $\BLFGSM$ with the contrastive loss
    \begin{equation*}
        \bar{\mathcal{L}}(f, X) = \sum_{i=1}^N - \log \frac{\exp(f(x_i)^\top f(\kappa(x_i))/ T)} {\sum_{j=1}^N \exp(f(x_j)^\top f(\kappa(x_i))/ T) }.
    \end{equation*}
    Here adversarial examples are selected to \emph{jointly} maximize the contrastive loss.

    \item The adversarial training of AdvCL generates adversarial attacks by maximizing a multi-view contrastive loss computed over the adversarial example, two views of $x$ and its high-frequency component $\highpass(x)$ \citep{fan_when_2021}.
    It corresponds to \BLPGD\ with the loss
    \begin{align*}
        \mathcal L (f,X,\hat X_u) &= \frac{1}{N} \sum_{i=1}^N \mathcal L' \left(\kappa_1(x_i), \kappa_2(x_i), \hat x_{i,u}, \highpass(x_i); f, X \right), \\
        \mathcal L'(z_1,\ldots, z_m; f, X) &= - \sum_{i=1}^m \sum_{\substack{j=1\\j\neq i}}^m \log \frac
        {\exp(\Sim (f(z_i), f(z_j)) / T ) }
        { \sum_{z_k\in X } \sum_{\kappa\in\{\kappa_1,\kappa_2\}} \exp(\Sim(f(z_i), f(\kappa(z_k)) )/T )  },
    \end{align*}
    with $\Sim(\cdot,\cdot)$ being the cosine similarity.

    \item Using the NT-Xent loss and the \BLPGD\ attack on a pair of views of $x$ is identical to the Adversarial-to-Adversarial and Dual Stream adversarial contrastive learning methods by \citet{jiang_robust_nodate}.

\end{itemize}

\section{Extended results}
\label{app:extended_results}

\begin{table}[p]
    \caption{Extended results for ResNet50-based models and adversarially fine-tuned MOCOv2. Arrows show if larger or smaller values are better.}
    \label{tab:extended_results_resnet}
    \centering
    \includegraphics[scale=0.55]{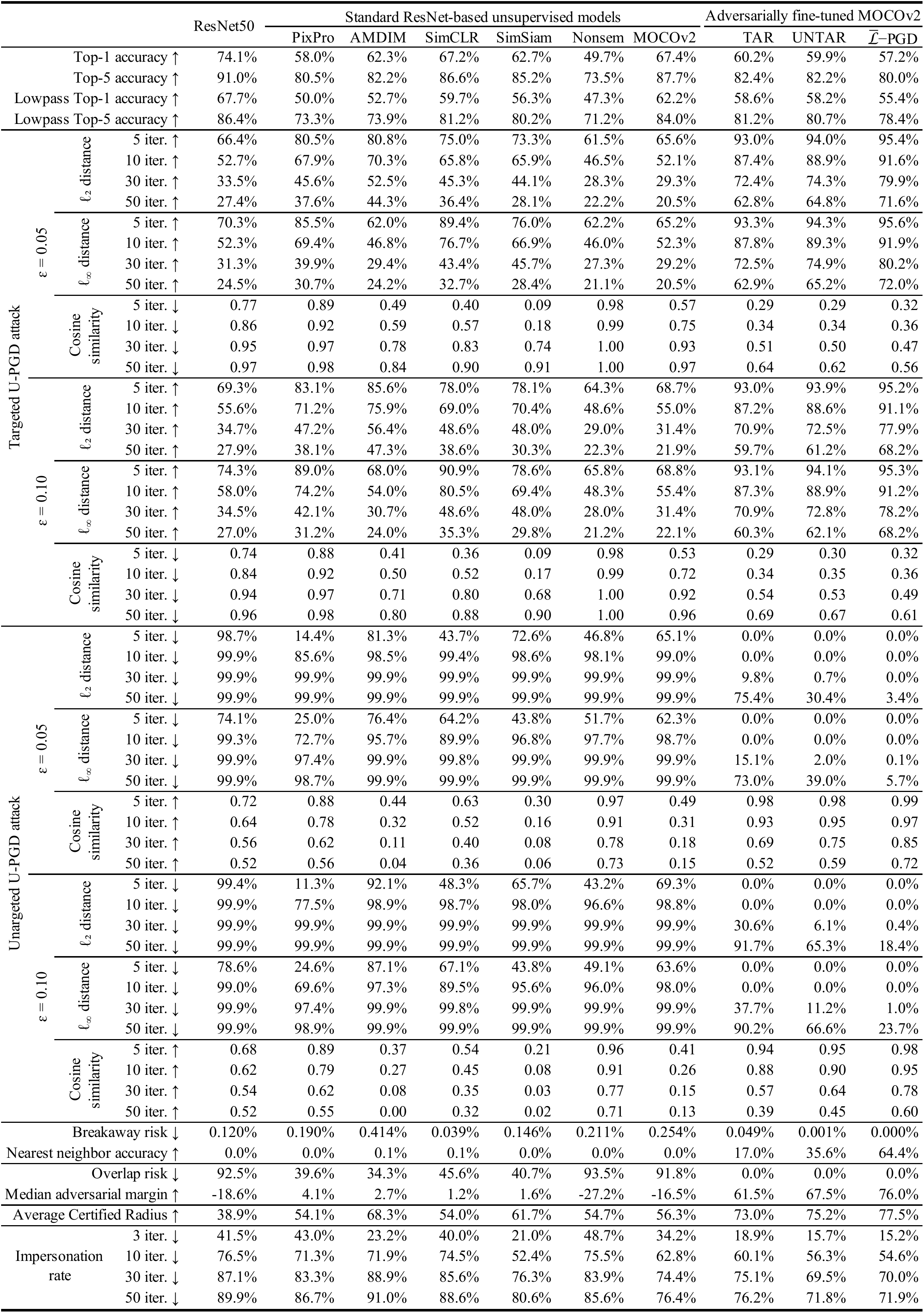}
\end{table}

\begin{table}[p]
    \caption{Extended results for transformer-based models and adversarially fine-tuned MOCOv3. Arrows show if larger or smaller values are better.}
    \label{tab:extended_results_transformer}
    \centering
    \includegraphics[scale=0.55]{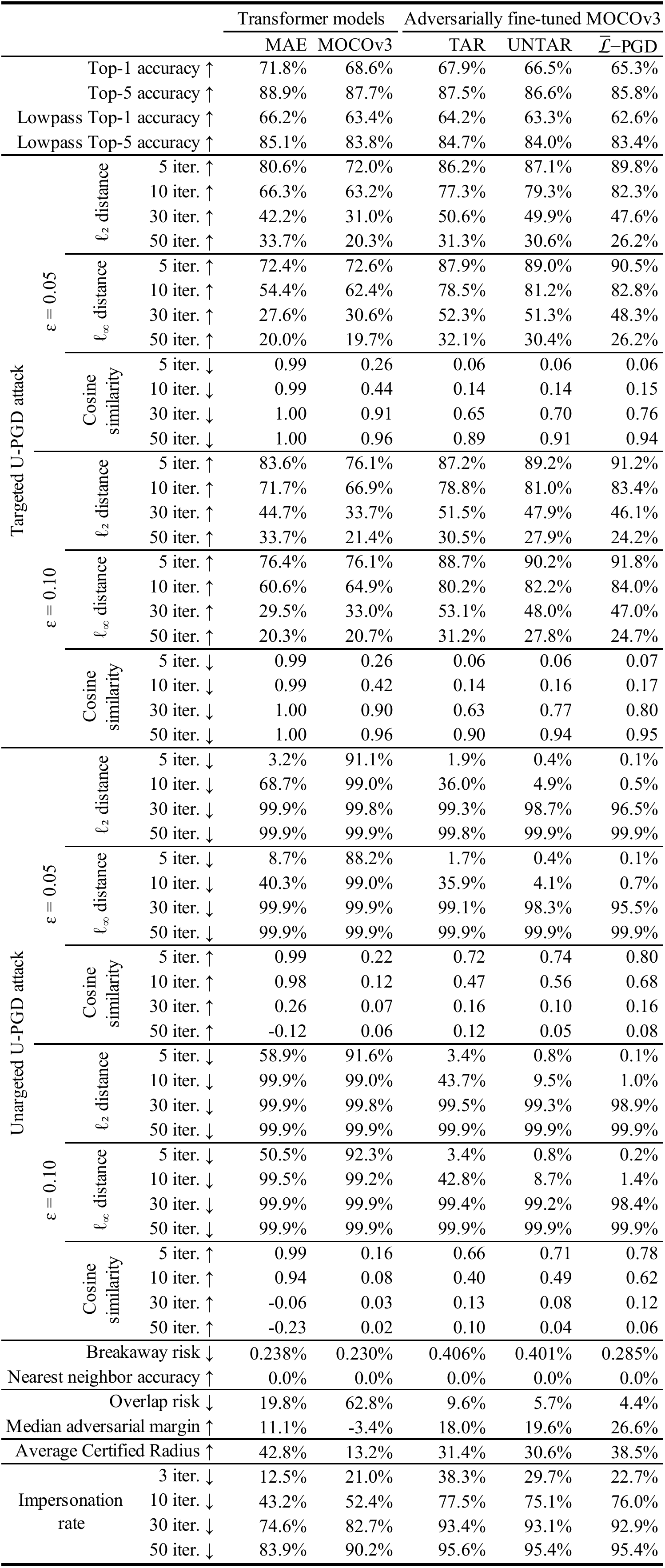}
\end{table}

\begin{figure}
    \centering
    \begin{minipage}[t]{0.48\textwidth}
        \centering
        \includegraphics[width=\textwidth]{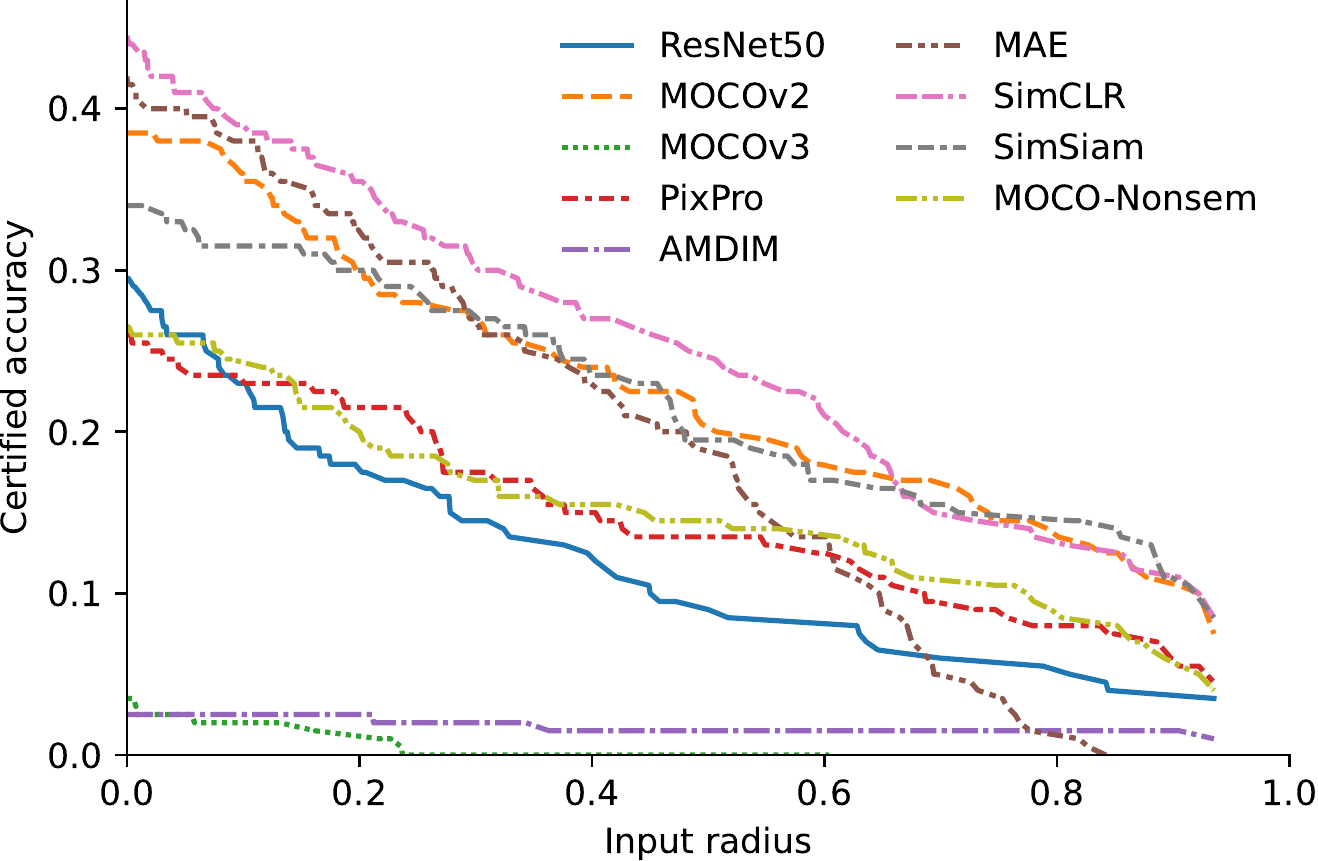}
        \caption{Certified accuracy of the standard models on ImageNet.}
        \label{fig:certified_accuracy_others}
    \end{minipage}\hfill
    \begin{minipage}[t]{0.48\textwidth}
        \centering
        \includegraphics[width=\textwidth]{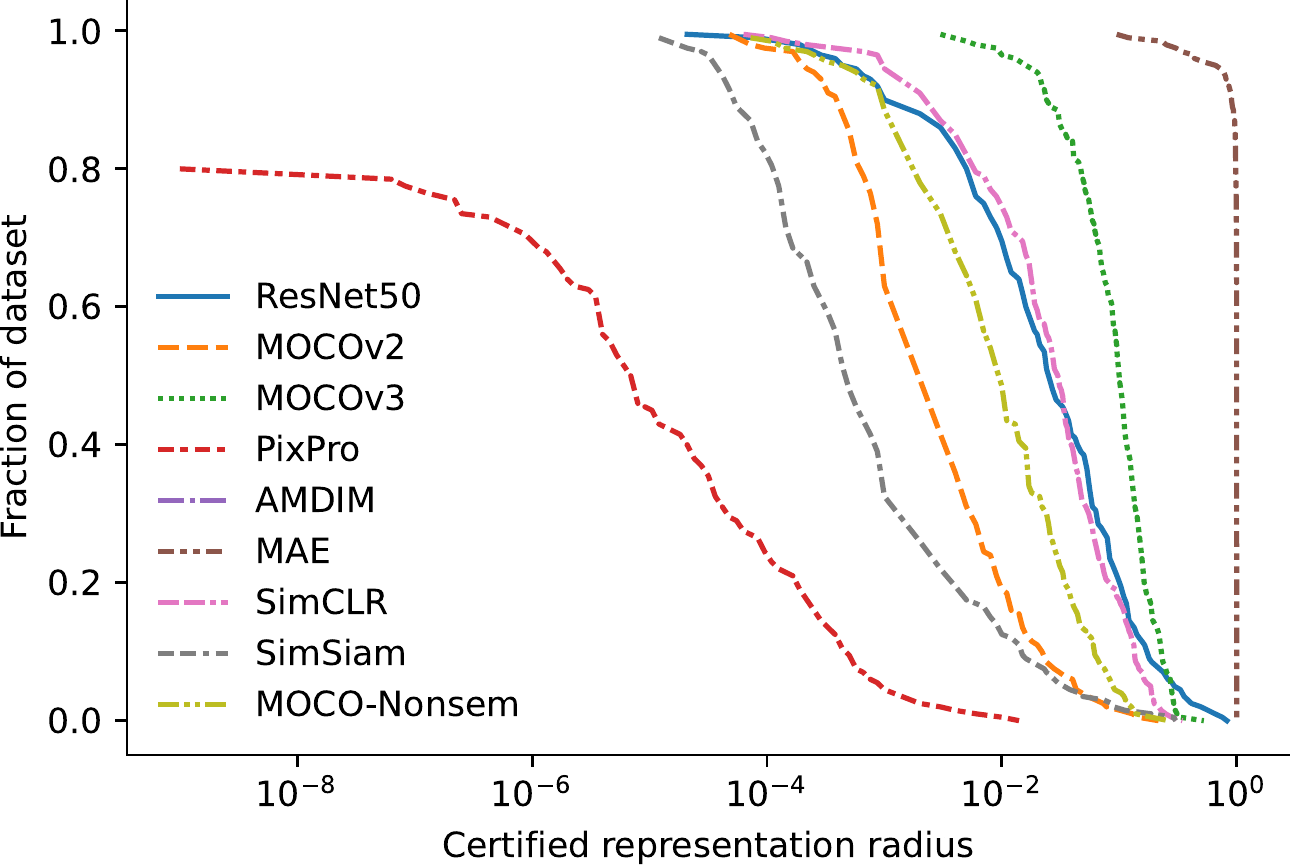}
        \caption{Certified robustness of the standard models on PASS using center smoothing. The distribution of certified radii in $\RR$ reported as percentile of the distribution of clean representation distances is shown. Smaller values indicate higher unsupervised robustness.}
        \label{fig:certified_robustness_others}
    \end{minipage}

    \vspace{10mm}
    \begin{minipage}[t]{0.48\textwidth}
        \centering
        \includegraphics[width=\textwidth]{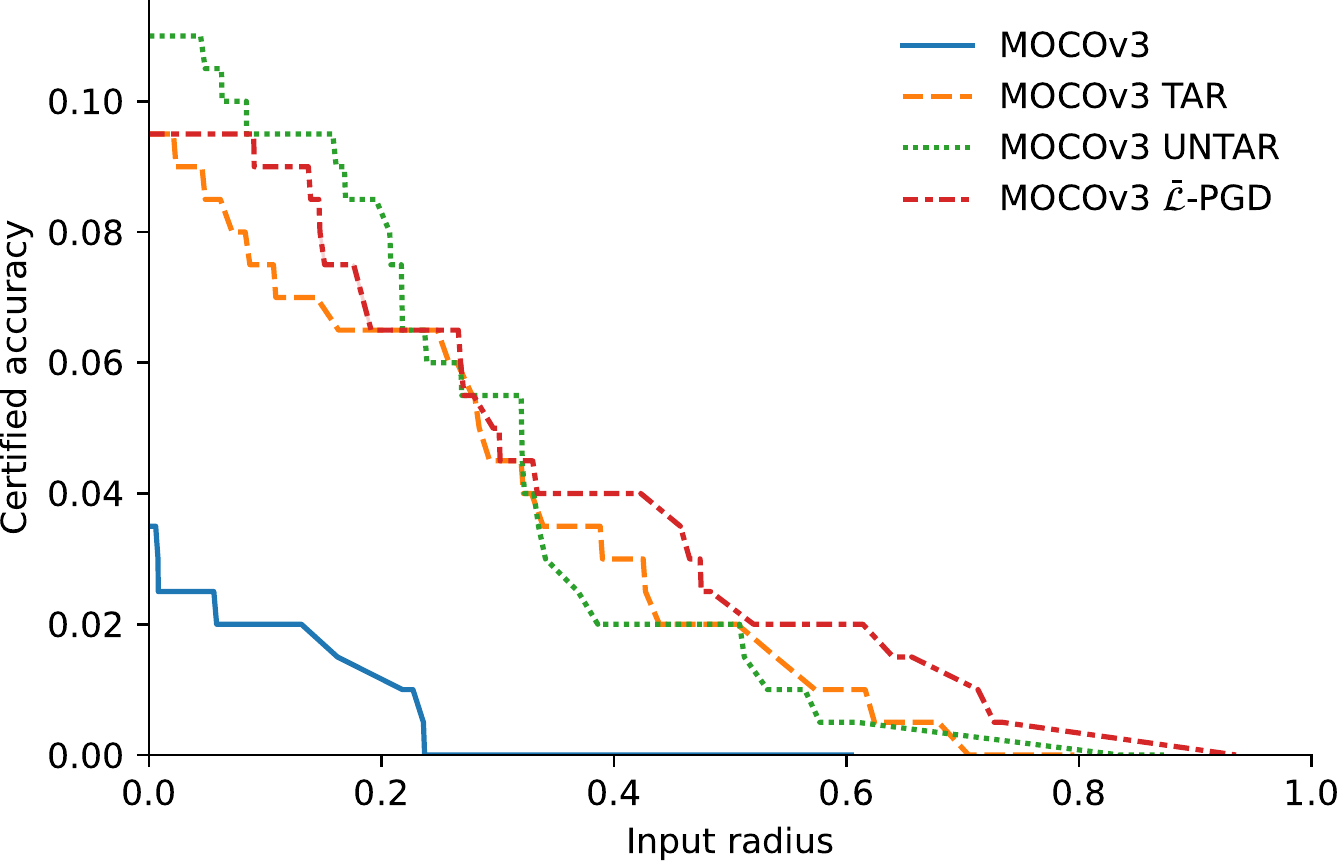}
        \caption{Certified accuracy of MOCOv3 and its adversarially trained variants on ImageNet computed via randomized smoothing. The adversarially trained models are uniformly certifiably more robust for almost all radius values.}
        \label{fig:certified_accuracy_moco3}
    \end{minipage}\hfill
    \begin{minipage}[t]{0.48\textwidth}
        \centering
        \includegraphics[width=\textwidth]{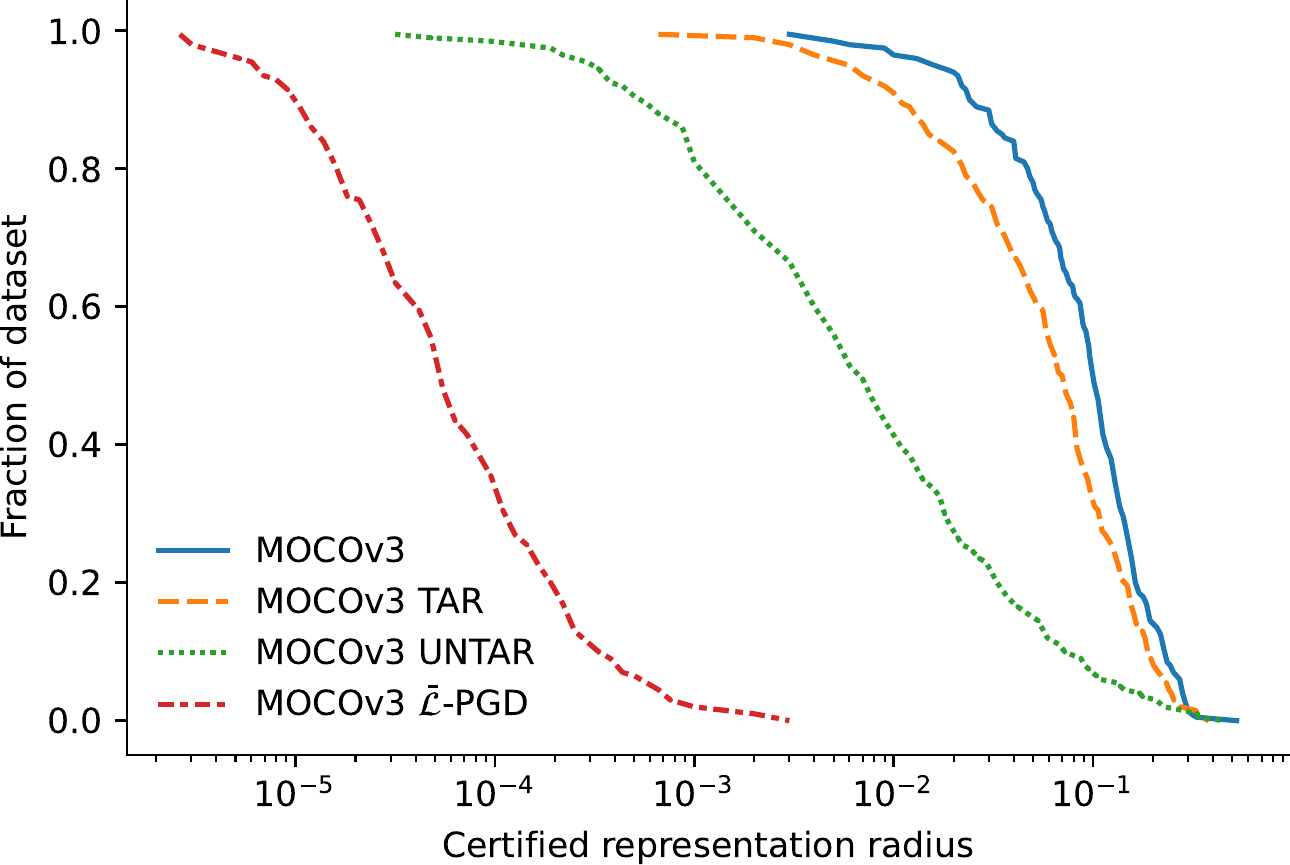}
        \caption{Certified robustness of MOCOv3 on PASS using center smoothing. The distribution of certified radii in $\RR$ reported as a percentile of the distribution of clean representation distances is shown. Smaller values indicate higher unsupervised robustness. }
        \label{fig:certified_robustness_moco3}
    \end{minipage}
\end{figure}

This appendix presents more comprehensive experimental results in \Cref{tab:extended_results_resnet,tab:extended_results_transformer,fig:certified_accuracy_others,fig:certified_robustness_others,fig:certified_accuracy_moco3,fig:certified_robustness_moco3}.

In addition to the ResNet50-based models discussed in \Cref{sec:experiments,sec:results}, we also present results from two models with transformer architectures \citep{vaswani_attention_2017}. 
MAE uses masked autoencoders \citep{he_masked_2021} and we use its ViT-Large variant.
MOCOv3 is a modification of MOCOv2 to work with a transformer backbone (ViT-Small) \citep{chen_empirical_2021}.
We apply the three unsupervised adversarial fine-tuning techniques from the main text to MOCOv3 to compare how they transfer to transformer architectures.
We use the exact same attack types and parameter values as for MOCOv2.

For all models, we present top-1 and top-5 accuracy for both the standard and the lowpass settings in \Cref{tab:extended_results_resnet,tab:extended_results_transformer}.
We report the results for the $\ell_2$- and $\ell_\infty$-induced divergence in representation space, at iterations 5, 10, 30 and 50 for U-PGD attacks with both $\epsilon=0.05$ and $\epsilon=0.10$. 
The attacks are performed with the $\ell_2$-induced divergence and $\alpha=0.001$.
These are reported as universal quantiles for the untargeted attacks and relative quantiles for the targeted attacks.
We also report the breakaway and overlap risks, as well as the nearest neighbor accuracy, median adversarial margin, average certified radius and impersonation rates as in the main text.
\Cref{fig:certified_robustness_others,fig:certified_accuracy_moco3,fig:certified_robustness_others,fig:certified_accuracy_others} show the certified robustness and accuracy of the models which were omitted from the main text.

Some of the models, including MOCOv2, are trained with a contrastive objective based on the cosine similarity.
Therefore, using the Euclidean distances as the divergence in representation space could be considered an unfair comparison as the adversarially trained models are optimized for it while the standard models are optimized for the cosine similarity.
Therefore, for targeted attacks, we also report the median cosine similarity between the representations of the adversarial examples and the representations of the target samples.
Higher values mean that the attack is more successful and that the model is less robust.
For the untargeted attacks, we report the median cosine similarity between the representations of the adversarial examples and the representations of the original samples.
Hence, higher values mean that the attack is less successful and the model is more robust.
The results in \Cref{tab:extended_results_resnet,tab:extended_results_transformer} show that adversarial training with the $\ell_2$-induced divergence leads also to improvements when measuring the cosine similarity: the divergence that the standard models are trained for but the adversarial ones are not.
This evidence supports our claim that the improvements we see from unsupervised adversarial fine-tuning are not due to our choice of divergence.

Amongst the standard models, MAE outperforms the ResNet50-based models on most measures. 
It has the highest accuracy of all models trained in the unsupervised regime, i.e.\ excluding the supervised ResNet50.
MAE also has some of the best robustness against targeted U-PGD attacks and is competitive to PixPro for the untargeted case.
It attains the lowest breakaway and overlap risks among all standard models as well as the largest median adversarial margin and is most robust to impersonation attacks.

MOCOv3 scores rather poorly in comparison across most measures though.
Hence, the robustness of MAE cannot be solely attributed to the transformer backbone.
This complements the observed variation in robustness performance among the ResNet50 models and provides further evidence that it is the objective function, rather than the backbone architecture, that determines robustness.

The three adversarially trained MOCOv3 models witness a lower accuracy penalty than the corresponding MOCOv2 models: between 0.2\% and 3.3\% for the clean and between -0.9\% and 0.8\% for the lowpass accuracy for MOCOv3 compared with correspondingly 5.3\%-10.2\% and 2.8\%-6.8\% for MOCOv2.
In fact, the adversarially trained MOCOv3 TAR model has a \emph{higher} lowpass accuracy than the standard MOCOv3.
Unsupervised adversarial training also leads to a uniformly better robustness against targeted and untargeted U-PGD attacks, albeit with a lower improvement compared to MOCOv2.
We similarly witness large improvements in the overlap risk and median adversarial margin measures.
The average certified radius and the certified accuracy (\Cref{fig:certified_accuracy_moco3}) are also significantly improved by adversarial training.
\Cref{fig:certified_robustness_moco3} shows that the adversarially trained models are also more certifiably robust than the baseline MOCOv3.

However, the three adversarially trained models actually \emph{fare worse} than the baseline MOCOv3 for breakaway risk and are \emph{less robust} to impersonation attacks.
The impersonation attacks are also less semantic in nature than the ones for the MOCOv2 adversarially trained models (\Cref{fig:impersonation_moco_tar,fig:impersonation_moco_untar,fig:impersonation_moco_loss} vs \Cref{fig:impersonation_moco3_tar,fig:impersonation_moco3_untar,fig:impersonation_moco3_loss}).
This could also be due to the learning rate being too low, rather than due to transformer models being inherently more difficult to adversarially fine-tune in an unsupervised setting.
The lower accuracy gap and the lower robustness further indicate that the adversarial training might not have been as ``aggressive'' for MOCOv3 as it was for MOCOv2.
Still, while the improvements for MOCOv3 are not as drastic as for MOCOv2, unsupervised adversarial training does improve most robustness measures.
The lower effectiveness for MOCOv3 of unsupervised adversarial training, especially in its role as a defence against impersonation attacks, is an avenue for future work that should examine whether there are fundamental differences between unsupervised adversarial training of CNN and Transformer models.

\section{Training and evaluation details}
\label{app:implementation_details}

This section provides further details on the unsupervised adversarial training and the evaluation metrics implementations.

\subsection{Unsupervised adversarial training for MOCOv2}
\label{app:moco2_training}

The three variants for the adversarially trained MOCOv2 are obtained by using a modification of the \href{https://github.com/facebookresearch/moco}{official MOCO source code}.
We perform fine-tuning by resuming the training procedure for additional 10 epochs but with the modified training loop.
The unsupervised adversarial examples are concatenated to the model's $q$-inputs and the $k$-inputs are correspondingly duplicated as shown in \Cref{algo:adversarial_training}.
For \BLPGD\ we use InfoNCE \citep{oord_representation_2019}, the loss that MOCOv2 is trained with.
All parameters, including the learning rate and its decay are as used for the original training and as reported by \citet{he_momentum_2020}.
We only reduced the batch size from 256 to 192 in order to be able to train on four GeForce RTX 2080 Ti GPUs.

\begin{lstlisting}[caption={Pseudocode of adversarial fine-tuning for MoCo (modified from \citep{he_momentum_2020}).},
  label={algo:adversarial_training},
  ]
# f_q, f_k: encoder networks for query and key
# queue: dictionary as a queue of K keys (CxK)
# m: momentum
# t: temperature

f_k.params = f_q.params # initialize

for x in loader: # load a minibatch x with N samples
    x_q = aug(x) # a randomly augmented version
    x_k = aug(x) # another randomly augmented version
    
    # perform the attack
    switch attack_type:
        case targeted:
            target_representation = roll(f_q(x_k), shifts=1)
            x_adv = targeted_upgd(x_q, target_representation)

        case untargeted:
            x_adv = untargeted_upgd(x_q)

        case loss:
            x_adv = batch_loss_upgd(f_q, x_q, f_k, x_k, queue, m, t)

    # get the representations
    q_clean = f_q.forward(x_q) # queries: NxC
    q_adv = f_q.forward(x_adv) # adversarial: NxC
    q = cat([q_clean, q_adv], dim=0)
    k = f_k.forward(x_k) # keys: NxC
    k = k.detach() # no gradient to keys

    # positive logits: 2Nx1
    l_pos = bmm(q.view(2*N,1,C), cat([k, k], dim=0).view(2*N,C,1))
    # negative logits: 2NxK
    l_neg = mm(q.view(2*N,C), queue.view(C,K))
    # logits: 2Nx(1+K)
    logits = cat([l_pos, l_neg], dim=1)
    # contrastive loss
    labels = zeros(2N) # positives are the 0-th
    loss = CrossEntropyLoss(logits/t, labels)
    # SGD update: query network
    loss.backward()
    update(f_q.params)
    # momentum update: key network
    f_k.params = m*f_k.params+(1-m)*f_q.params
    # update dictionary
    enqueue(queue, k) # enqueue the current minibatch
    dequeue(queue) # dequeue the earliest minibatch
\end{lstlisting}

\subsection{Unsupervised adversarial training for MOCOv3}
\label{app:moco3_training}

Similarly to MOCOv2, the three variants for the adversarially trained MOCOv3 are obtained by using a modification of the \href{https://github.com/facebookresearch/moco-v3}{official MOCOv3 source code}.
We perform fine-tuning by resuming the training procedure for additional 10 epochs but with the modified training loop.
The unsupervised adversarial examples are added to the contrastive loss as shown in \Cref{algo:adversarial_training_moco3}.
All parameters, are as used for the original training and as reported by \citet{chen_empirical_2021}.
We only increased the learning rate from $1.5\times 10^{-4}$ to $1.5\times 10^{-3}$ and reduced the batch size from 256 to 192 in order to be able to train on four GeForce RTX 2080 Ti GPUs.

\begin{lstlisting}[caption={Pseudocode of adversarial fune-tuning for MOCOv3.},
  label={algo:adversarial_training_moco3},
  ]
# f_base: base encoder network
# f_predictor: predictor network
# m: momentum

f_momentum.params = f_q.params # initialize momentum encoder

for x in loader: # load a minibatch x with N samples
    x_0 = aug(x) # a randomly augmented version
    x_1 = aug(x) # another randomly augmented version
    
    # perform the attack
    switch attack_type:
        case targeted:
            target_representation = roll(f_predictor(f_base(x_1)), shifts=1)
            x_adv = targeted_upgd(x_0, target_representation)

        case untargeted:
            x_adv = untargeted_upgd(x_0)

        case loss:
            x_adv = batch_loss_upgd(f_base, f_predictor, x_0, x_1)

    # update the momentum encoder
    f_momentum.params = f_momentum.params * m + f_base.params * (1-m)

    # get the base representations
    q_0 = f_predictor.forward(f_base.forward(x_0))     # x_0 reps: NxC
    q_1 = f_predictor.forward(f_base.forward(x_1))     # x_1 reps: NxC
    q_adv = f_predictor.forward(f_base.forward(x_adv)) # attacked reps: NxC

    # get the momentum representations
    k_0 = f_momentum.forward(x_0)     # x_0 reps: NxC
    k_1 = f_momentum.forward(x_1)     # x_1 reps: NxC
    k_adv = f_momentum.forward(x_adv) # attacked reps: NxC


    # compute the loss
    loss = contrastive_loss(q_0, k_1) + contrastive_loss(q_1, k_0) + \
           contrastive_loss(q_adv, k_1) + contrastive_loss(q_1, k_adv)

    # parameter update: query network
    loss.backward()
    update(f_q.params)
\end{lstlisting}

\subsection{Linear probes}
\label{app:linear_probes}
As part of the evaluation we train three linear probes for each model.
\begin{itemize}
    \item \textbf{Standard linear probe:} for computing the top-1 and top-5 accuracy on clean samples, as well as for the impersonation attack evaluation.
    \item \textbf{Lowpass linear probe:} for computing the top-1 and top-5 accuracy on samples with removed high-frequency components. We use the implementation of \citet{wang_high-frequency_2020} and keep only the Fourier components that are within a radius of 50 from the center of the Fourier-transformed image.
    \item \textbf{Gaussian noise linear probe:} trained on samples with added Gaussian noise for computing the certified accuracy as randomized smoothing results in a more robust model when the base model is trained with aggressive Gaussian noise \citep{Lecuyer_certified_2019}. Therefore, we add Gaussian noise with $\sigma=0.25$ to all inputs.
\end{itemize}

All linear probes are trained with the train set of ImageNet Large Scale Visual Recognition Challenge 2012 \citep{russakovsky_imagenet_2015} and are evaluated on its test set.
For training we use modification of the \href{https://github.com/facebookresearch/moco/blob/main/main_lincls.py}{MOCO linear probe evaluation code} for 25 epochs.
The starting learning rate is 30.0 with 10-fold reductions applied at epochs 15 and 20.

For fairness of the comparison, we use the same implementation to evaluate all models.
Therefore, there might be differences between the accuracy values reported by us and the ones reported in the original publications of the respective models.

\subsection{Computing the distribution of inter-representational divergences}
The distribution of $\ell_2$ and $\ell_\infty$-induced divergences between the representations of clean samples of PASS \citep{asano_pass_2021} is needed for computing the universal and relative quantiles.
Due to computational restrictions, we compute the representations of 10,000 samples and the divergences between all pairs of them in order to construct the empirical estimate of the distribution of inter-representational divergences.
We observe that 10,000 samples are more than sufficient for the empirical estimate of the distribution to converge.

\subsection{Adversarial attacks}

In \Cref{tab:results_resnet_robustness,tab:results_advtrained_resnet_robustness,tab:extended_results_resnet,tab:extended_results_transformer} we report U-PGD attacks performed with $d(x,x')=\left\|x-x'\right\|_2$ and $\alpha=0.001$.
We report median values over the same 1000 samples from PASS \citep{asano_pass_2021}.
The median universal quantile is reported for targeted attacks and the median relative quantile is reported for targeted attacks as explained in \Cref{sec:assessing_robustness}.
In \Cref{tab:results_resnet_robustness,tab:results_advtrained_resnet_robustness} we report only the resulting $\ell_2$ quantiles, while in the extended results (\Cref{tab:extended_results_resnet,tab:extended_results_transformer}) we also show the $\ell_\infty$ quantiles and cosine similarities.

\subsection{Breakaway risk and nearest neighbor accuracy}
\label{app:breakaway_impl_details}
The breakaway risk and nearest neighbor accuracy are also computed by attacking the same 1000 samples from PASS ($D'$) and computing their divergences with all other samples from PASS ($D$).
Our empirical estimate is then :
$$  \hat p_\text{breakaway} = \frac{1}{|D'|(|D|-1)} \sum_{x_i \in D'} \sum_{x_j \in D/\{x_i\}}  \mathbbm{1} \left[ d(f(\hat x_i), f(x_j)) < d( f(\hat x_i), f(x_i)) \right], $$
where $\hat x_i$ is the untargeted U-PGD attack with $d(x,x')=\left\|x-x'\right\|_2$, $\epsilon=0.05$ and $\alpha=0.001$ for 25 iterations.

\subsection{Overlap risk and median adversarial margin}
\label{app:overlap_impl_details}
The overlap risk and median adversarial margin are computed over 1000 pairs  of samples from PASS ($D'$).
Each element of the pair is attacked to have a representation similar to the other element.
$$  \hat p_\text{overlap} = \frac{1}{|D'|} \sum_{(x, x') \in D'} \mathbbm{1} \left[ d(f(x_i), f(\hat{x}_j^{\to i})) < d(f(x_i), f(\hat{x}_i^{\to j})) \right], $$
where $\hat x_i^{\to j}$ is the targeted U-PGD attack on $x_i$ towards $x_j$ with $d(x,x')=\left\|x-x'\right\|_2$, $\epsilon=0.05$ and $\alpha=0.001$ for 10 iterations.

\subsection{Certified accuracy}
\label{app:certified_accuracy}

We use the  \href{https://github.com/locuslab/smoothing}{randomized smoothing implementation} by \citet{cohen_certied_2019}.
We evaluate the Gaussian noise linear probe (see \Cref{app:linear_probes}) over 200 samples from the ImageNet test set \citep{russakovsky_imagenet_2015}.
We use $\sigma=0.25, N_0=100, N=100,000$ and an error probability $\alpha=0.001$, as originally used by \citet{cohen_certied_2019}.
\Cref{fig:certified_accuracy,fig:certified_accuracy_moco3} show the resulting certified accuracy for MOCOv2, MOCOv3, and their unsupervised adversarially trained versions.
\Cref{fig:certified_accuracy_others} shows the certified accuracy for the other models.
These plots show the fraction of samples which are correctly classified and which certifiably have the same classification within a given $\ell_2$ radius of the input space.

\Cref{tab:extended_results_resnet,tab:extended_results_transformer} also show the Average Certified Radius for all models.
The Average Certified Radius was proposed by \citet{Zhai2020MACER} as a way to summarize the certified accuracy vs radius plots with a single number representing the average certified radius for the correctly classified samples.

\subsection{Certified robustness}
\label{app:certified_robustness}

The certified robustness evaluation in \Cref{fig:certified_robustness,fig:certified_accuracy_moco3} was done with the \href{https://github.com/aounon/center-smoothing}{center smoothing implementation} by \citet{kumar_center_2021}.
We evaluate the models over the same 200 samples from the ImageNet test set \citep{russakovsky_imagenet_2015}.
We use $\sigma=0.25, N_0=10,000, N=100,000$ and error probabilities $\alpha_1=0.005,\alpha_2=0.005$, as originally proposed by \citet{kumar_center_2021}.

\subsection{Impersonation attacks}
\label{app:impersonation_attacks_impl_details}
The impersonation attack evaluation is performed using targeted U-PGD attacks.
We use the Assira dataset that contains 25,000 images, equally split between cats and dogs \citep{elson_asirra_2007}.
When evaluating a model, we consider only the subset of images that the standard linear probe (see \Cref{app:linear_probes}) for the given model classifies correctly.
Then, we construct pairs of an image of a cat and an image of a dog. 
We perform two attacks: attacking the cat to have a representation as close as possible to that of the dog and vice-versa.
The attacked images are then classified with the clean linear probe.
The success rate of cats impersonating dogs and dogs impersonating cats are computed separately and then averaged to account for possible class-based differences.
Note that the linear probe is \emph{not} used for constructing the attack, i.e.\ we indeed fool it without accessing it.
\Cref{app:impersonation_attacks_samples} shows examples of the impersonation attacks for all models.

\section{Impersonation attacks}
\label{app:impersonation_attacks_samples}

This appendix showcases samples of the impersonation attacks on the models discussed in the paper.
The first and third row in each sample are the original images of cats and dogs respectively.
The second row is the result when each cat image is attacked to have a representation close to the representation of the corresponding dog image.
The fourth row is the opposite: the dog image attacked to have a representation close to the representation of the cat image.
The attack used was targeted U-PGD with $d(r,r')=\left\|r-r'\right\|_2$ for 50 iterations with $\epsilon=0.10$ and $\alpha=0.01$.
The samples shown differ from model to model as we restrict the evaluation to the samples that are correctly predicted by the given model, see \Cref{app:impersonation_attacks_impl_details} for details.

\begin{figure}[h]
    \centering
    \includegraphics[width=.99\textwidth]{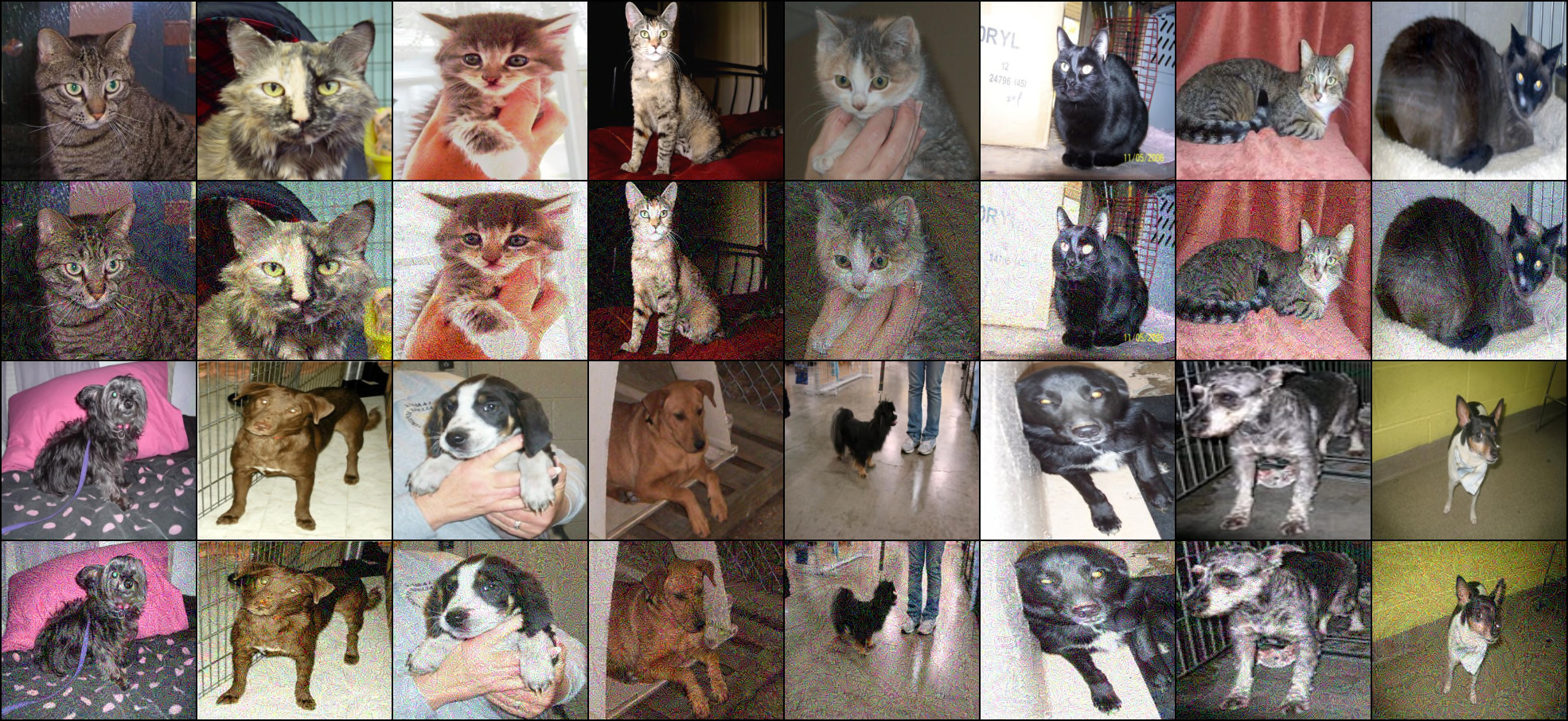}
    \caption{Impersonation attack samples for ResNet50 \citep{he_deep_2016}.}
    \label{fig:impersonation_resnet}
\end{figure}
\begin{figure}[h]
    \centering
    \includegraphics[width=.99\textwidth]{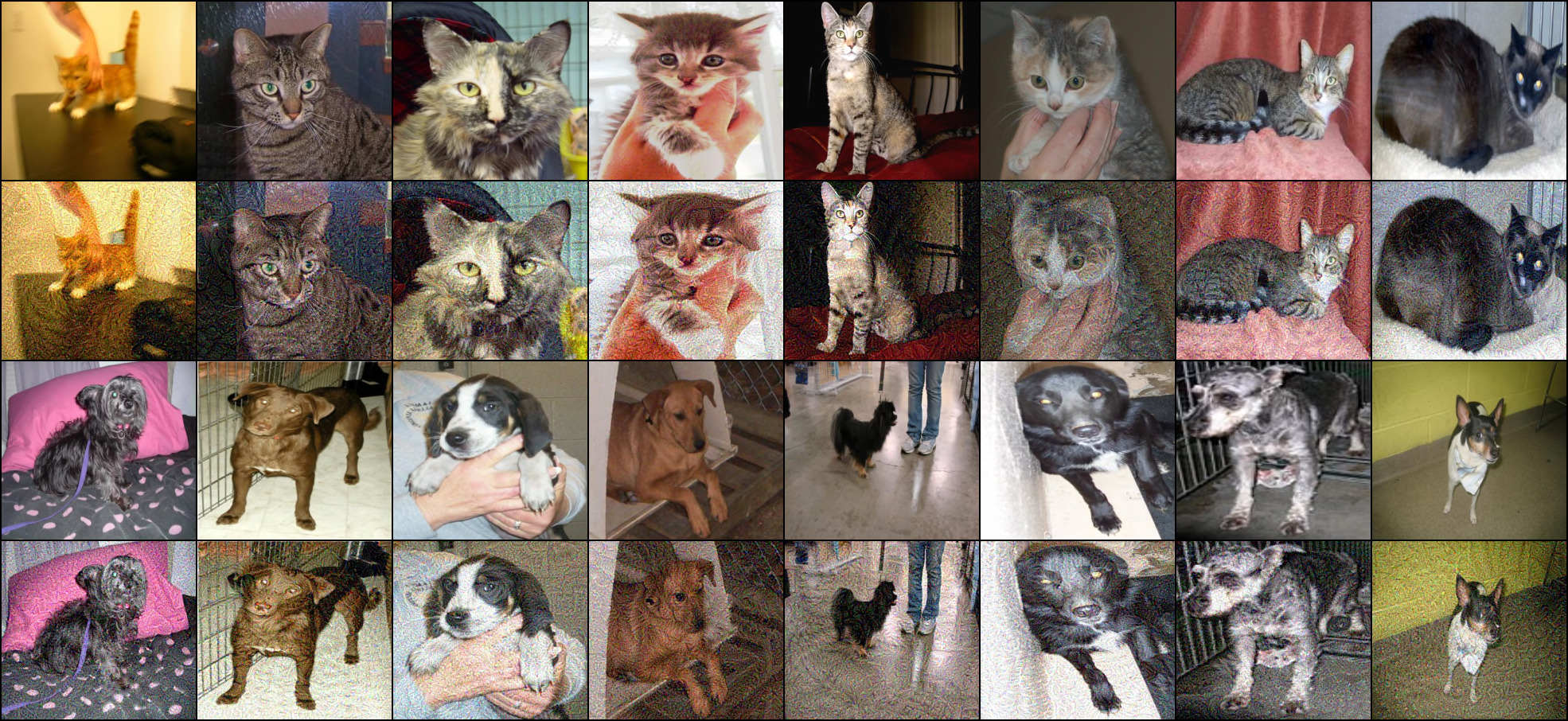}
    \caption{Impersonation attack samples for MOCO with non-semantic negatives \citep{ge_robust_nodate}.}
    \label{fig:impersonation_mocononsem}
\end{figure}
\begin{figure}[h]
    \centering
    \includegraphics[width=.99\textwidth]{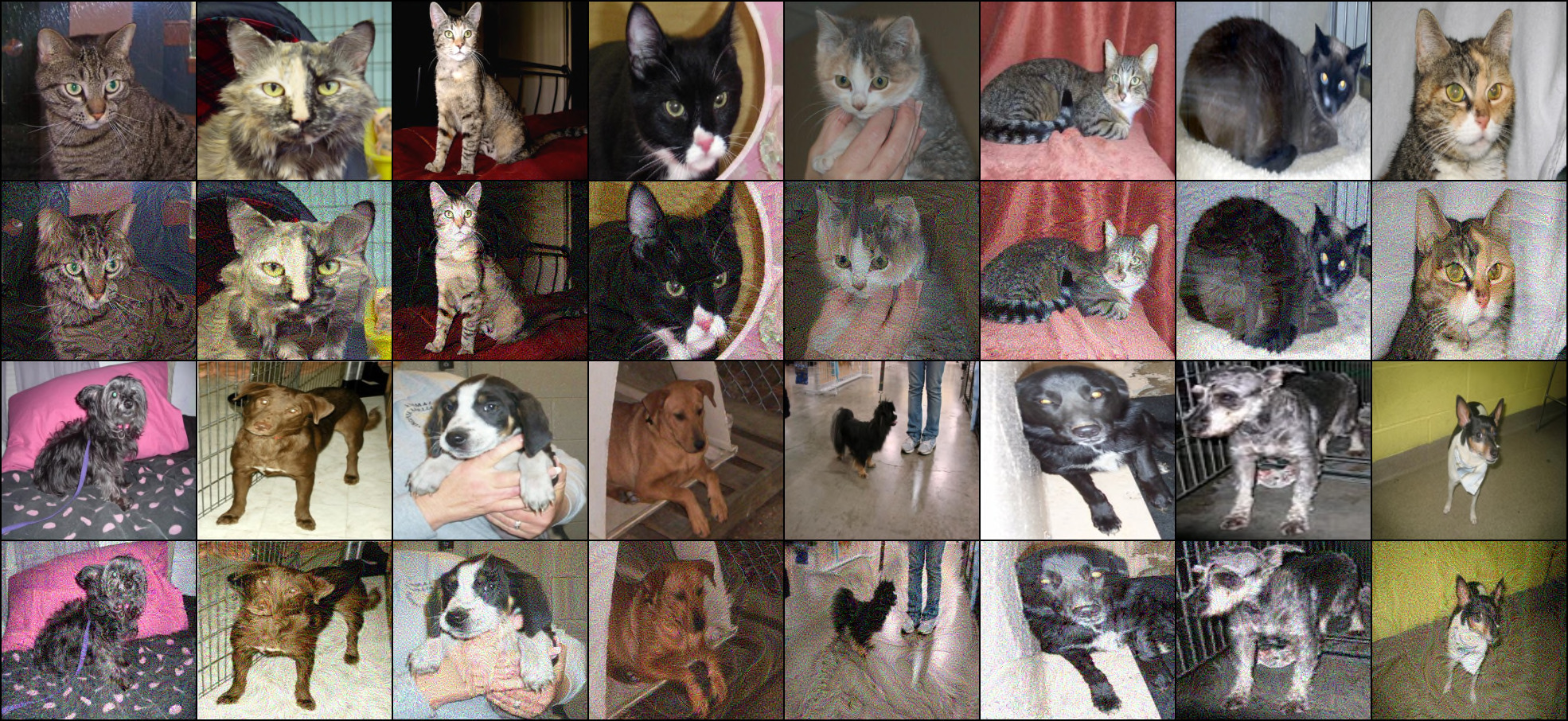}
    \caption{Impersonation attack samples for PixPro \citep{xie_propagate_2021}.}
    \label{fig:impersonation_pixpro}
\end{figure}
\begin{figure}[h]
    \centering
    \includegraphics[width=.99\textwidth]{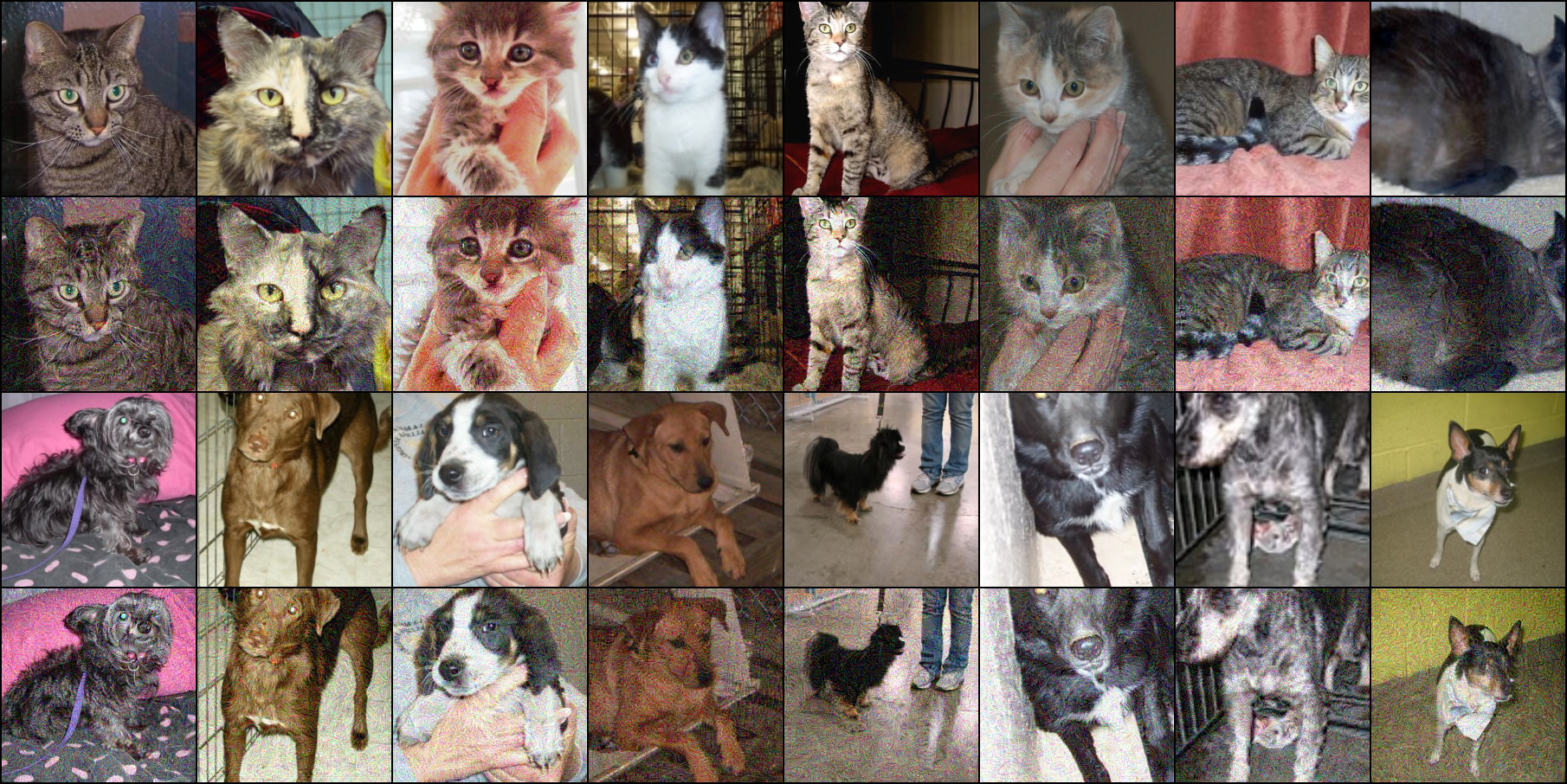}
    \caption{Impersonation attack samples for SimCLRv2 \citep{chen_big_nodate}.}
    \label{fig:impersonation_simclr}
\end{figure}
\begin{figure}[h]
    \centering
    \includegraphics[width=.99\textwidth]{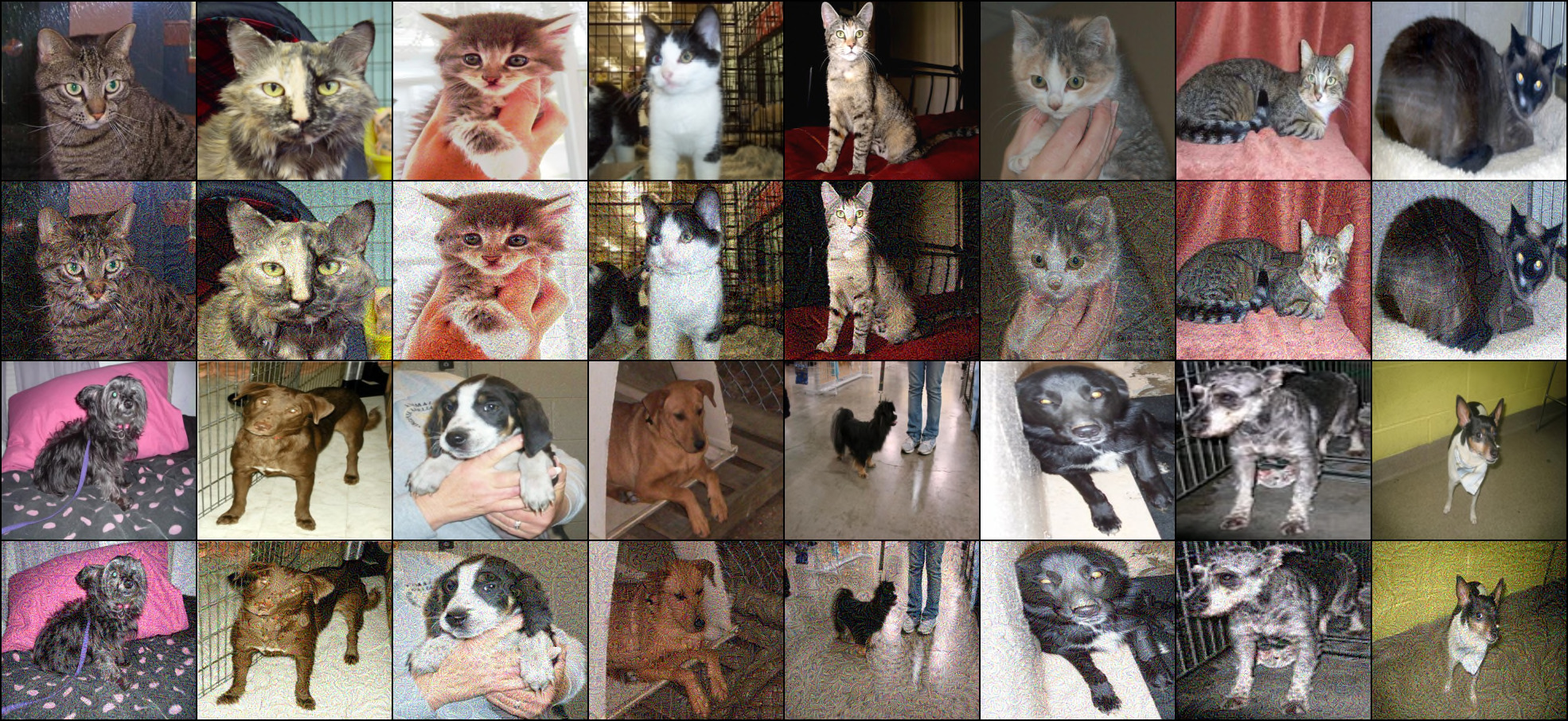}
    \caption{Impersonation attack samples for SimSiam \citep{chen_exploring_2021}.}
    \label{fig:impersonation_simsiam}
\end{figure}
\begin{figure}[h]
    \centering
    \includegraphics[width=.99\textwidth]{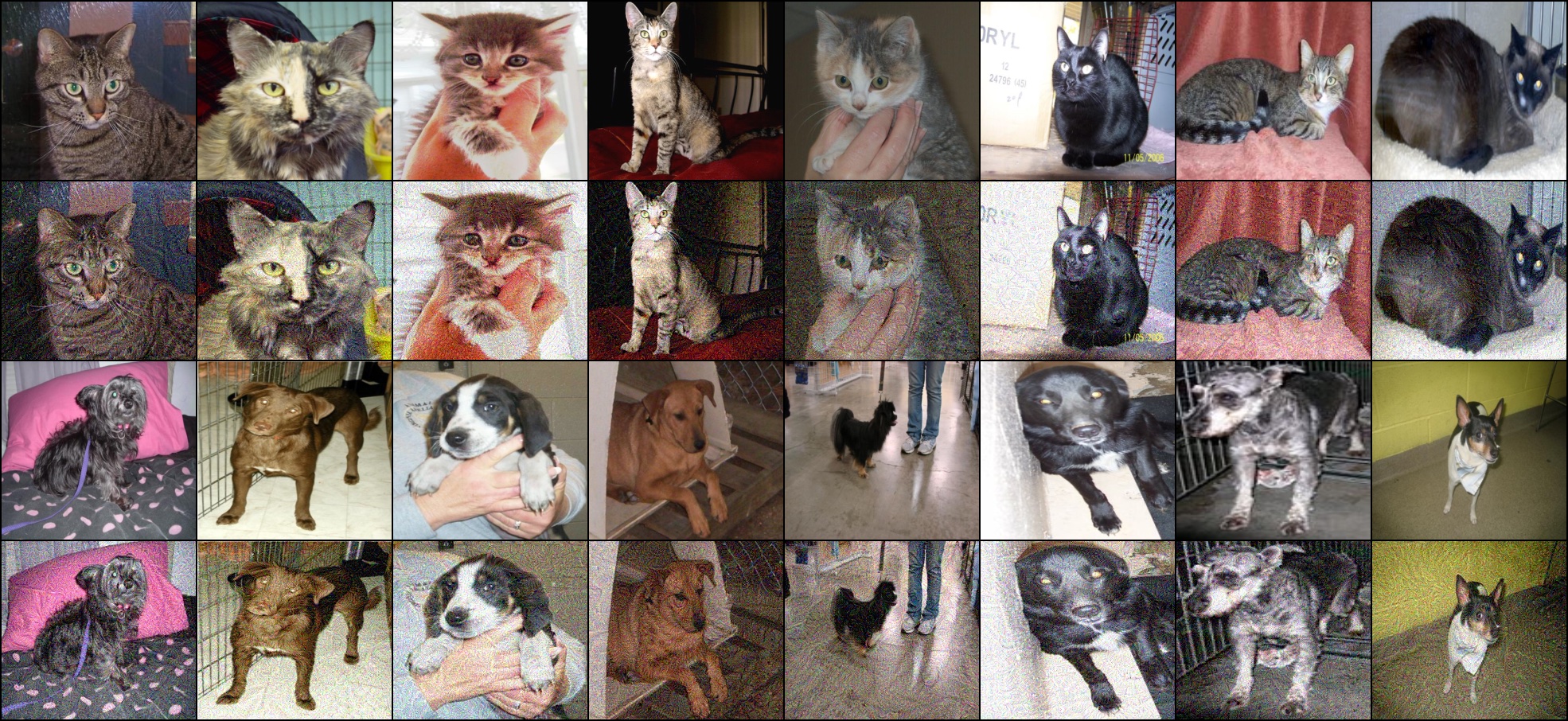}
    \caption{Impersonation attack samples for MOCOv2 \citep{he_momentum_2020,chen_improved_2020}.}
    \label{fig:impersonation_moco}
\end{figure}
\begin{figure}[h]
    \centering
    \includegraphics[width=.99\textwidth]{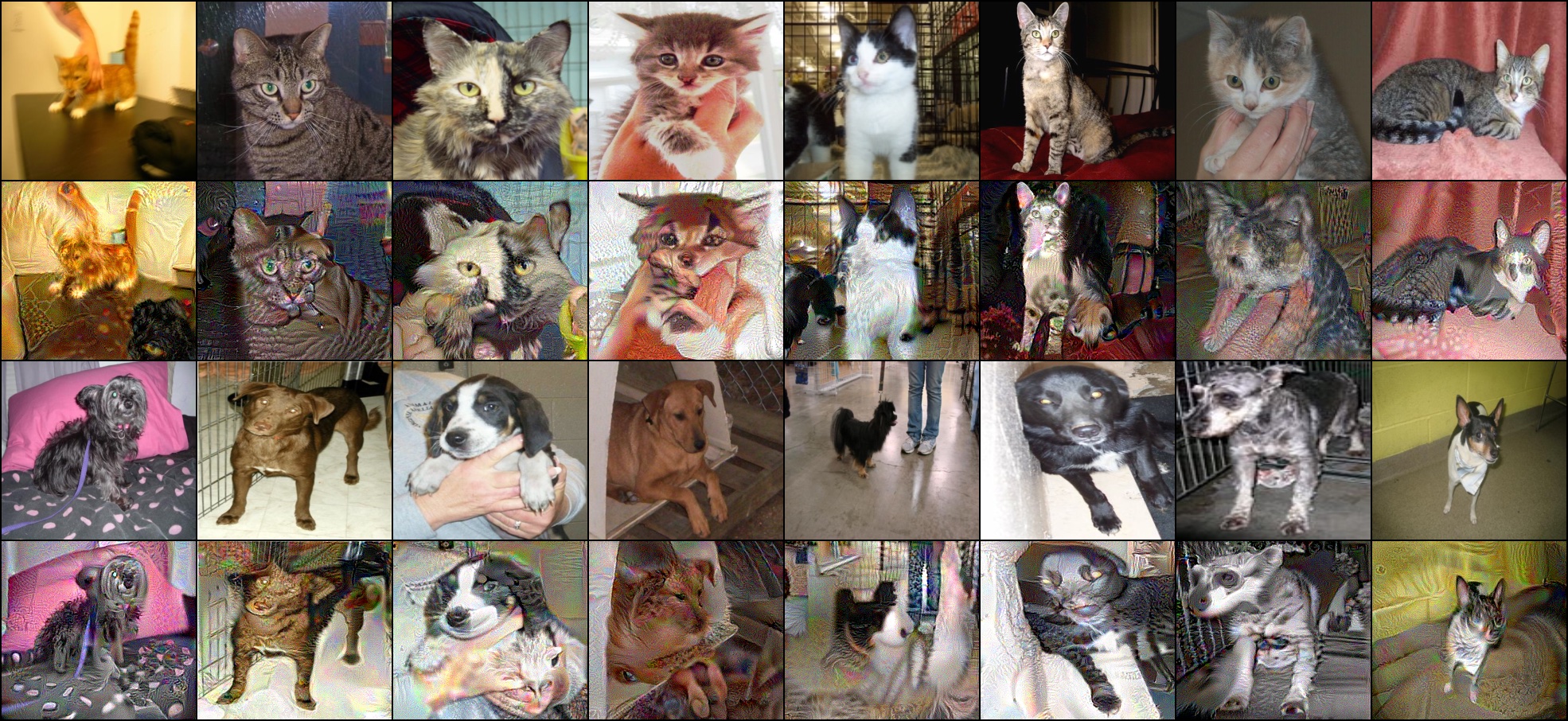}
    \caption{Impersonation attack samples for MOCOv2 TAR.}
    \label{fig:impersonation_moco_tar}
\end{figure}
\begin{figure}[h]
    \centering
    \includegraphics[width=.99\textwidth]{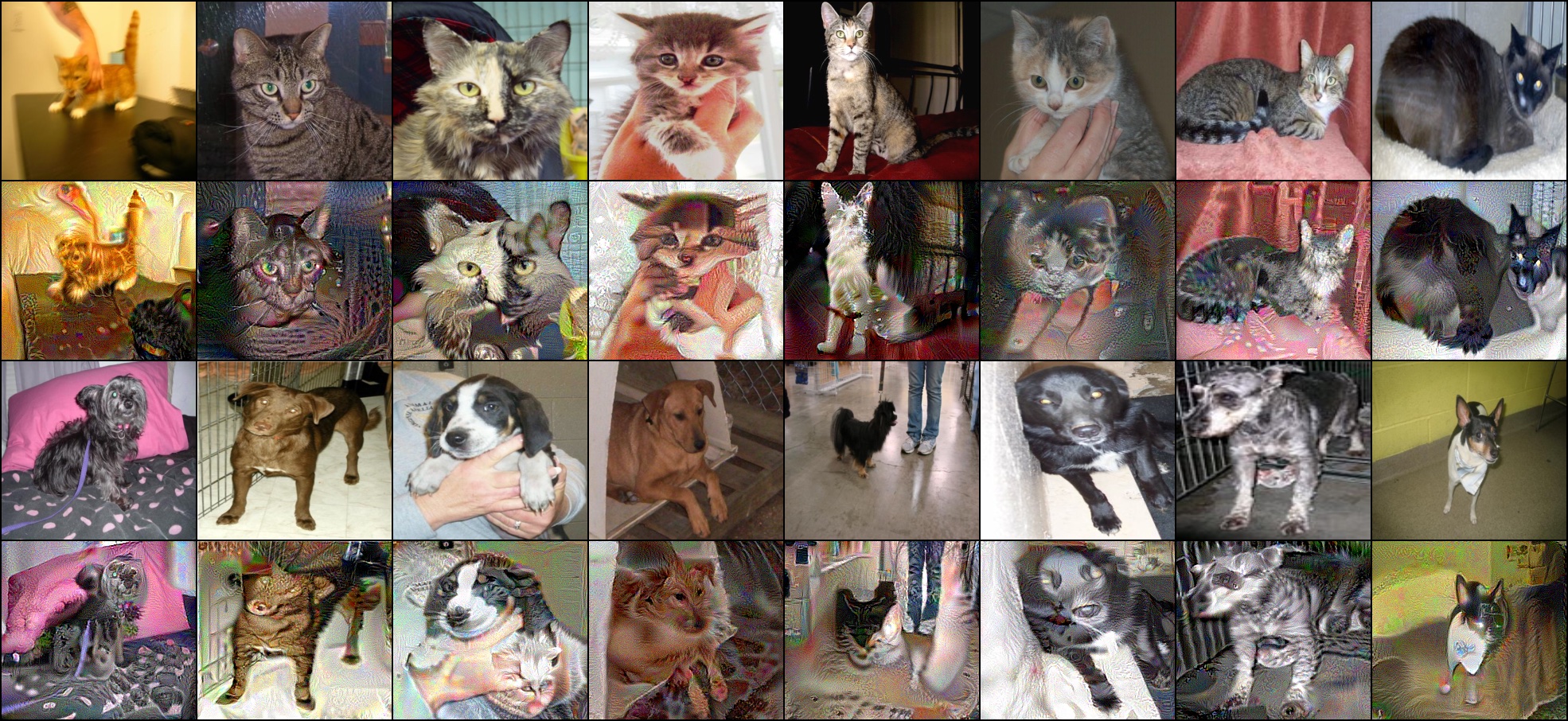}
    \caption{Impersonation attack samples for MOCOv2 UNTAR.}
    \label{fig:impersonation_moco_untar}
\end{figure}
\begin{figure}[h]
    \centering
    \includegraphics[width=.99\textwidth]{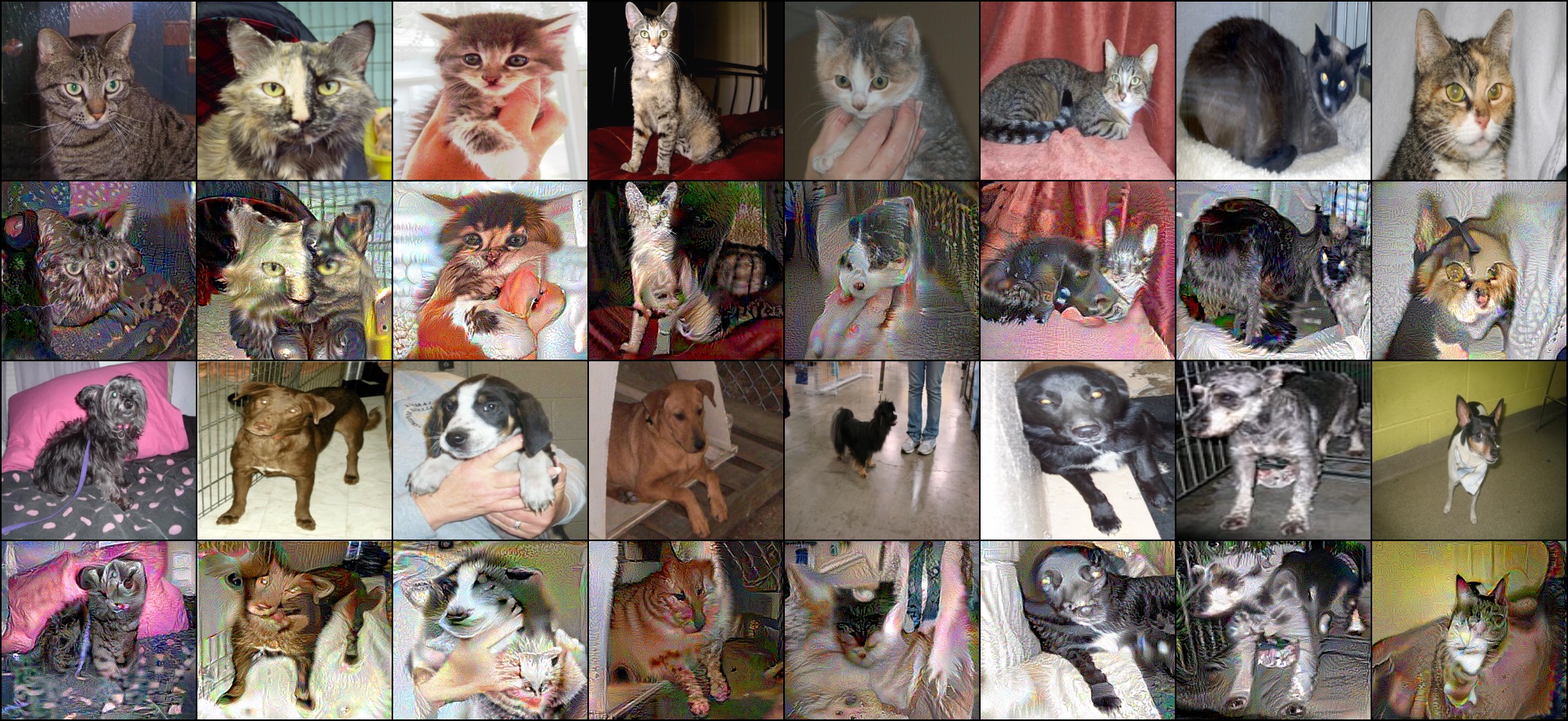}
    \caption{Impersonation attack samples for MOCOv2 LOSS.}
    \label{fig:impersonation_moco_loss}
\end{figure}
\begin{figure}[h]
    \centering
    \includegraphics[width=.99\textwidth]{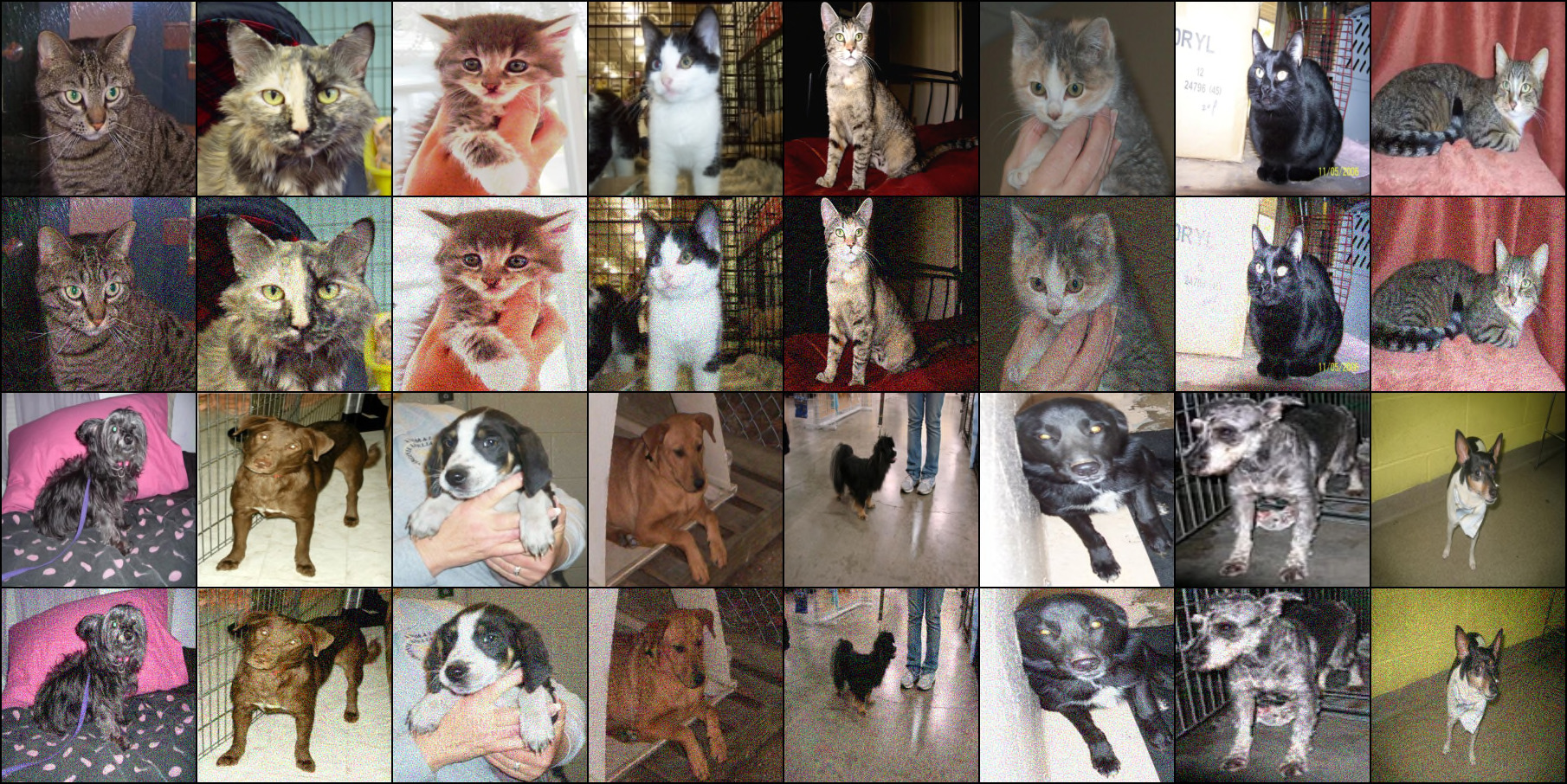}
    \caption{Impersonation attack samples for MOCOv3 \citep{chen_empirical_2021}.}
    \label{fig:impersonation_moco3}
\end{figure}
\begin{figure}[h]
    \centering
    \includegraphics[width=.99\textwidth]{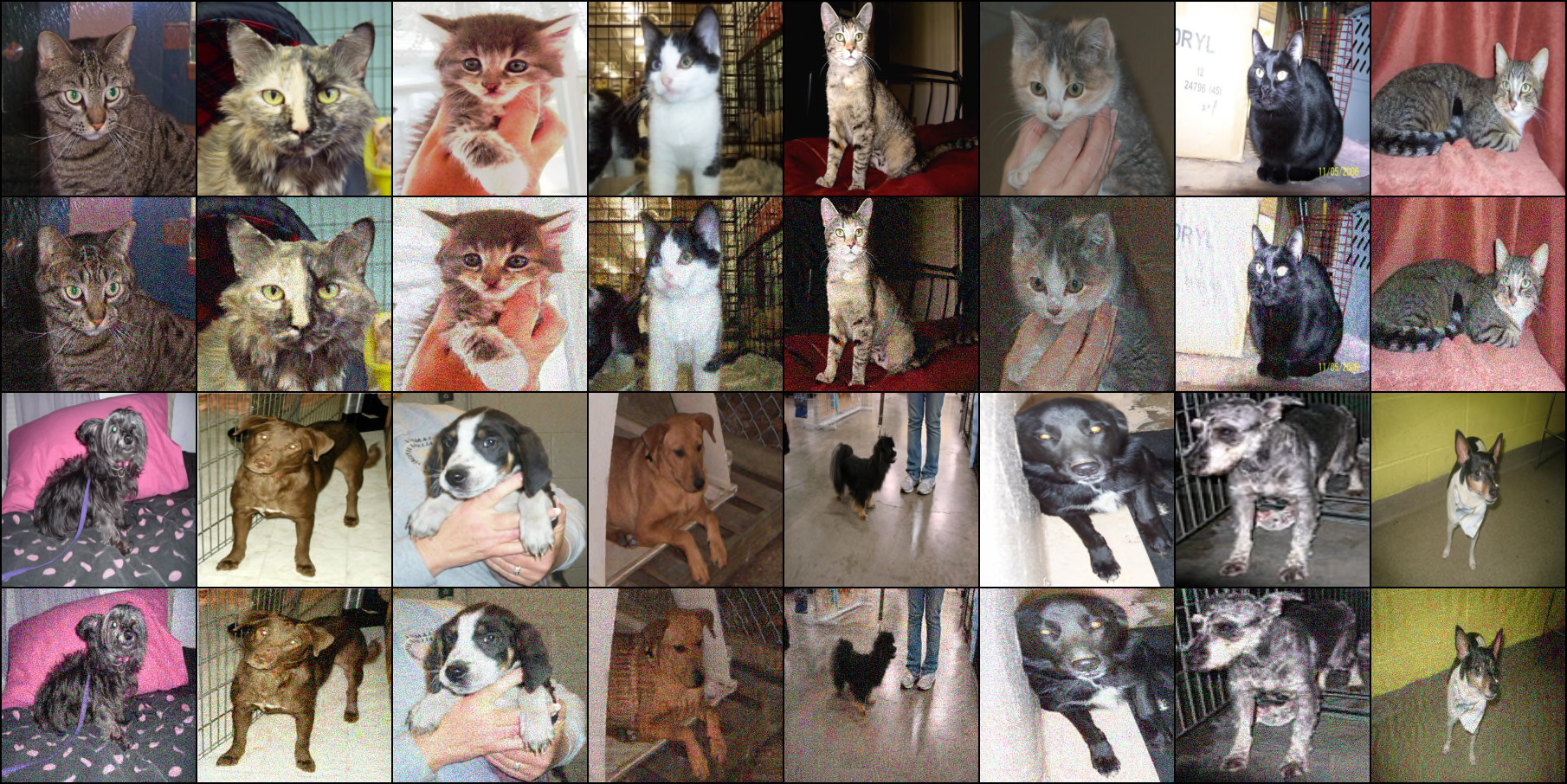}
    \caption{Impersonation attack samples for MOCOv3 TAR.}
    \label{fig:impersonation_moco3_tar}
\end{figure}
\begin{figure}[h]
    \centering
    \includegraphics[width=.99\textwidth]{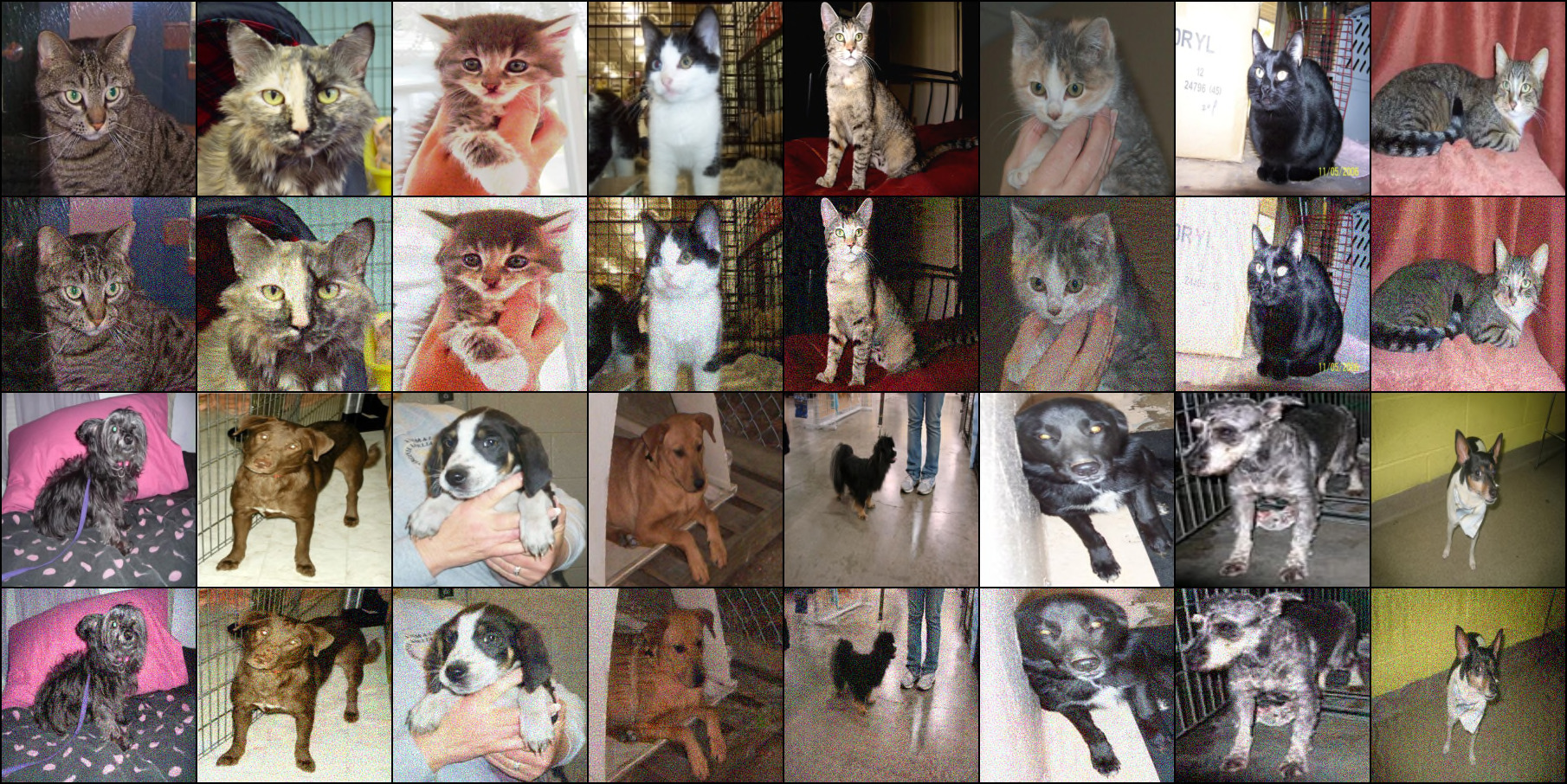}
    \caption{Impersonation attack samples for MOCOv3 UNTAR.}
    \label{fig:impersonation_moco3_untar}
\end{figure}
\begin{figure}[h]
    \centering
    \includegraphics[width=.99\textwidth]{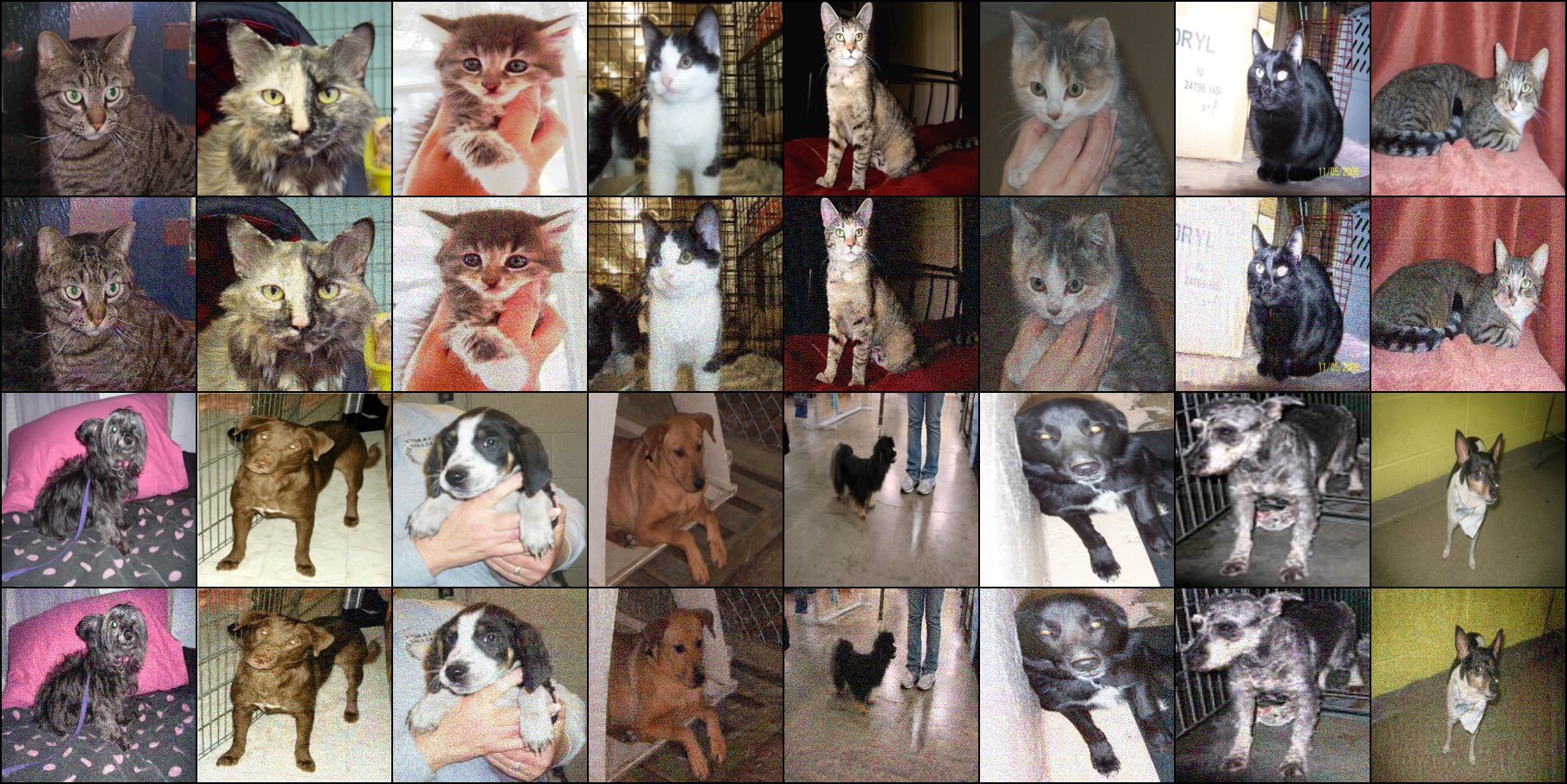}
    \caption{Impersonation attack samples for MOCOv3 LOSS.}
    \label{fig:impersonation_moco3_loss}
\end{figure}

A key observation is that the perturbations necessary to fool the standard models visually appear as noise (\Cref{fig:impersonation_resnet,fig:impersonation_moco,fig:impersonation_mocononsem,fig:impersonation_pixpro,fig:impersonation_simclr,fig:impersonation_simsiam}).
However, the perturbations applied to the three adversarially trained MOCOv2 models (\Cref{fig:impersonation_moco_tar,fig:impersonation_moco_untar,fig:impersonation_moco_loss}) are more ``semantic'' in nature, and in some cases even resemble features of the target class.
Still, this is not the case when comparing the impersonation attacks on MOCOv3 (\Cref{fig:impersonation_moco3}) with the attacks on the adversarially trained versions (\Cref{fig:impersonation_moco3_tar,fig:impersonation_moco3_untar,fig:impersonation_moco3_loss}).

\begin{figure}[h]
    \centering
    \includegraphics[width=.99\textwidth]{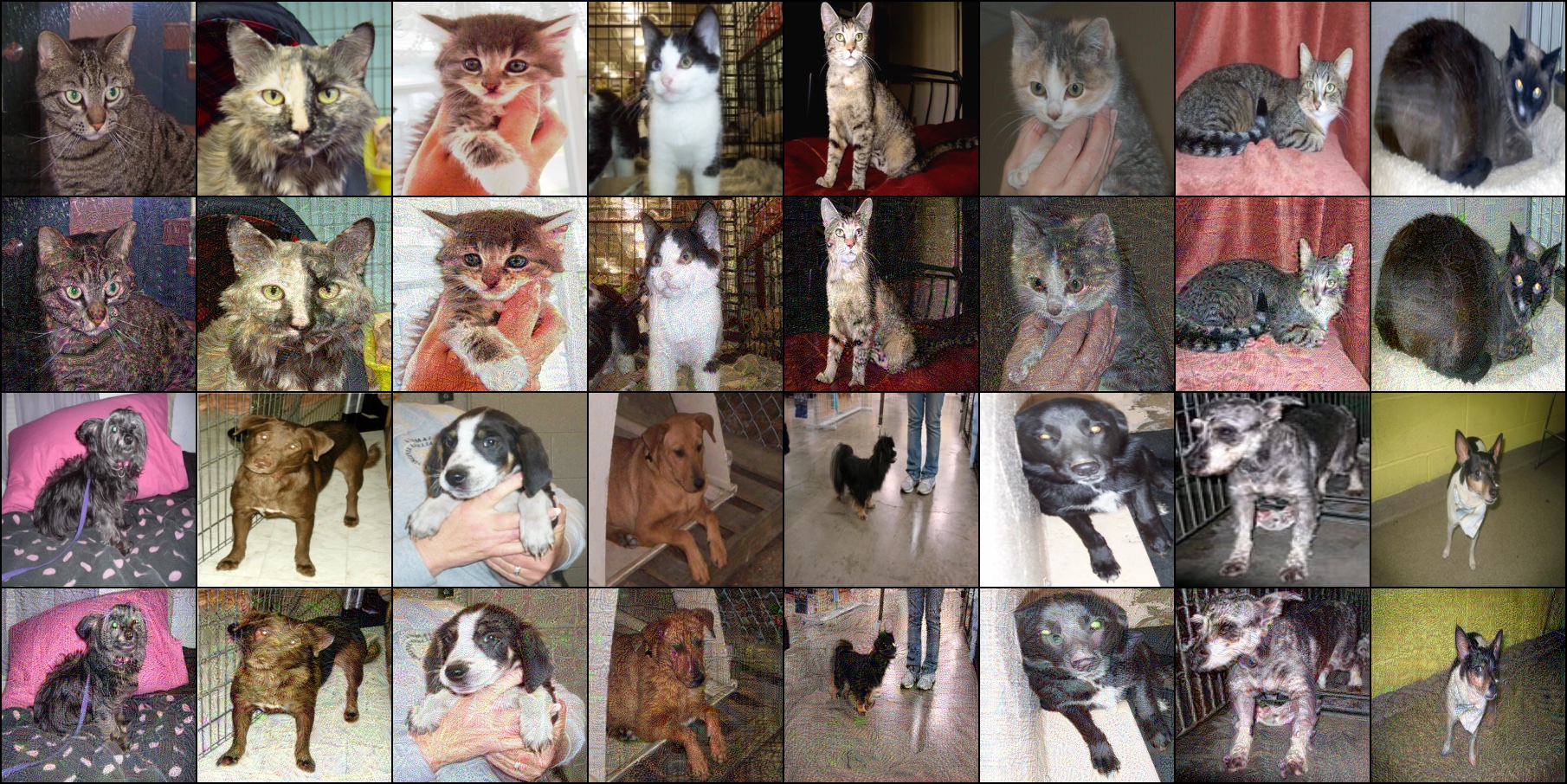}
    \caption{Impersonation attack samples for MAE \citep{he_masked_2021}.}
    \label{fig:impersonation_mae}
\end{figure}

\end{document}